\documentclass[journal,onecolumn,12pt, peerreview]{IEEEtran}
\usepackage{amsmath,amssymb}        
\usepackage{subcaption}
\usepackage{tabularx}
\usepackage{graphicx}
\usepackage{amsmath}
\usepackage{booktabs}
\usepackage{tabularx}
\usepackage{caption}
\usepackage[table]{xcolor}
\usepackage{placeins}    
\usepackage{hyperref}    
\usepackage{url} 
\usepackage{enumerate}
\usepackage{multirow} 
\usepackage{longtable}
\usepackage{makecell}
\usepackage{enumitem}

\newcommand{\ie}{\textit{i.e.}}
\newcommand{\eg}{\textit{e.g.}}
\newcommand{\etal}{\textit{et al.}}
\usepackage{doi} 

\begin{document}

\title{UStyle: Waterbody Style Transfer of Underwater Scenes by Depth-Guided Feature  Synthesis}

\author{Md Abu Bakr Siddique$^\star$, Vaishnav Ramesh$^\star$, Junliang Liu, Piyush Singh, \\ and Md Jahidul Islam
\\
{
\small RoboPI Laboratory, Department of ECE, University of Florida} 
\thanks{\noindent\rule{8cm}{0.6pt}}
\thanks{$^\star$~These two authors contributed equally}
\thanks{This pre-print is accepted for publication in the IEEE Journal of Oceanic Engineering (JOE).} \\
\thanks{Project page: \url{https://robopi.ece.ufl.edu/ustyle.html}} 
}



\maketitle

\begin{abstract}
 The concept of ``\textit{waterbody style}" transfer remains largely unexplored in the underwater imaging literature. Traditional image style transfer (STx) methods emphasize artistic and photorealistic blending that fails to preserve geometry in high-scattering underwater environments. The wavelength-dependent attenuation and depth-dependent backscattering of underwater optics further complicate STx learning from unpaired data. We introduce UStyle, the first data-driven framework for transferring waterbody styles across underwater images without needing reference images or explicit scene information. We propose a novel depth-aware whitening and coloring transform (\textbf{DA-WCT}) that incorporates physics-based waterbody synthesis for perceptually consistent stylization while preserving scene structure. We also integrate carefully designed loss functions to maintain color, lightness, structural integrity, frequency-domain features, and high-level content in VGG and CLIP spaces. Comprehensive experimental analyses show that UStyle surpasses state-of-the-art methods that rely solely on reconstruction loss. Additionally, we present the \textbf{UF7D dataset}, a curated benchmark of high-resolution underwater images across seven waterbody styles. The UStyle code and UF7D dataset are available at: \url{https://github.com/uf-robopi/UStyle}.
\end{abstract}

\begin{IEEEkeywords}
\vspace{-1mm}
Style Transfer; Waterbody Fusion; Underwater Imaging; Underwater Vision.
\end{IEEEkeywords}
\vspace{-5.0mm}
\section{Introduction}
Image style transfer (STx) in underwater imagery has significant potential in data augmentation, robotic vision, photometry, and imaging technologies~\cite{siddique2024aquafuse}. Traditional neural style transfer (NSTx) approaches~\cite{chen2024upst} primarily focus on terrestrial imagery, aiming to transfer artistic or photometric features across images. However, the concept of ``\textbf{waterbody style}" transfer presents a distinct challenge~\cite{siddique2024aquafuse}, where the goal is to generate an underwater image in the style of a different aquatic environment while preserving the structural integrity of the scene. Although unpaired images from various waterbodies can be collected, conventional NSTx techniques—optimized for perceptual photorealism~\cite{chiu2022photowct2} or artistic blending~\cite{chen2021artistic,deng2021arbitrary,deng2022stytr2}—struggle to accommodate the unique distortions introduced by underwater light absorption and scattering~\cite{akkaynak2018revised}.

\begin{figure*}[t]
    \centering
    \includegraphics[width=0.95\textwidth]{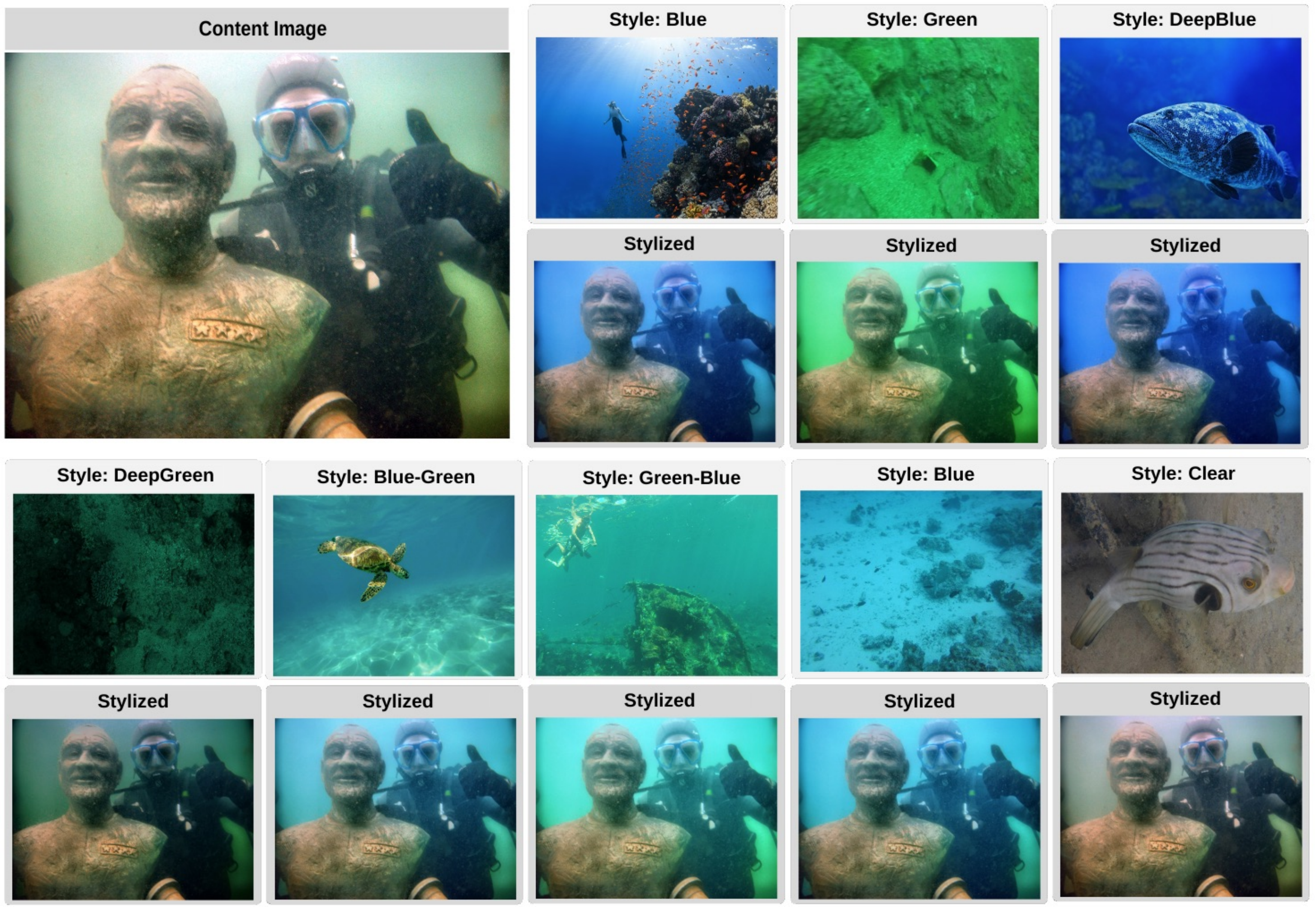}%
    \vspace{-1mm}
    \caption{Visualizations of UStyle's underwater STx capabilities are shown. The content image (top left) is transformed into various \textit{waterbody styles} to transfer the waterbody characteristics while preserving object geometry and structural details. 
    }
    \vspace{-2mm}
    \label{fig:intro}
\end{figure*}

Currently, no learning-based frameworks exist for waterbody STx in underwater imagery. Our previous work, AquaFuse~\cite{siddique2024aquafuse}, introduced a physics-based approach for \textit{waterbody fusion}, facilitating data augmentation across different aquatic environments. However, this method relies on a closed-form formulation that requires a carefully selected, noise-free reference image and prior scene information, including depth, incident angle, and attenuation parameters. Leveraging large-scale vision-based learning models for robust underwater STx without reference images or prior information remains a challenge. Recent advancements in NST~\cite{chiu2022photowct2,deng2022stytr2} and vision-language models~\cite{kwon2022clipstyler,yang2023zero} present an opportunity to address this gap.

In this paper, we propose \textbf{UStyle}, the first data-driven learning pipeline for waterbody STx in underwater imagery. We present the model architecture, choice of loss functions, and the training strategy to address the limitations of existing NSTx pipelines. UStyle integrates a residual encoder-decoder to extract local structural details at different spatial resolutions while leveraging progressive multi-stage training to refine waterbody features by domain-specific loss functions. To guide the stylization process, we use a physics-based method to synthesize the waterbody from style image by following the state-of-the-art (SOTA) underwater image formation model~\cite{akkaynak2018revised}. This synthesized \textit{waterbody style} is then fused by a novel depth-guided whitening and coloring transform (\textbf{DA-WCT}) mechanism, which makes sure that no object texture or scene information is transferred to the content image. A few sample examples are illustrated in Fig.~\ref{fig:intro}.

For comprehensive evaluation, we compile an underwater dataset named \textbf{UF7D} -- featuring $4,050$ high-resolution images with paired depth annotations across seven different waterbody styles: Blue (B), Green (G), Deep Blue (DB), Deep Green (DG), Clear (C), Blue-Green (B-G), and Green-Blue (G-B). It enables data-driven learning for underwater STx, allowing the model to generalize across diverse aquatic environments with varying optical properties. About $58$\% of the images are sourced from existing SOTA databases~\cite{li2019underwater,akkaynak2019sea, peng2023u,islam2020fast} for imaging, object detection, image enhancement, and robotics applications; thus, it serves as a useful benchmark for future research.  The remaining $42\%$ data are collected from a diverse set of open-access sources, where we solely focused on ensuring diversity of object-scenes geometry and waterbody types. We also release a \textbf{UF7D-Challenge} test set with $70$ samples across categories for more challenging evaluation of underwater STx methods. To this end, we conducted a thorough benchmark evaluation of underwater STx, by comparing UStyle with several learning-based as well as physics-based SOTA methods named 
StyTR-2~\cite{deng2022stytr2}, MCCNet~\cite{wang2020arbitrary}, ASTMAN~\cite{deng2020arbitrary}, IEContraAST~\cite{chen2021artistic}, StyleID~\cite{chung2024style}, PhotoWCT2~\cite{chiu2022photowct2}, and AquaFuse~\cite{siddique2024aquafuse}.

\vspace{1mm}
\noindent
The key contributions of this work are summarized below:
\begin{enumerate}[label={$\arabic*$)},nolistsep,leftmargin=*]
\item \textbf{Underwater STx learning pipeline and dataset:} UStyle is the first data-driven STx framework for underwater scenes, eliminating the need for explicit scene priors required by physics-based STx models~\cite{siddique2024aquafuse}. Instead of relying on carefully selected reference images and waterbody parameters, UStyle employs a learning-based STx approach that robustly adapts to diverse aquatic environments with varying waterbodies. We also introduce the UF7D dataset, a comprehensive collection of $4,050$ high-resolution underwater images of waterbody styles, which supports supervised training and benchmark evaluation across varying underwater conditions.

\item \textbf{Novel contributions in UStyle:} We design and develop UStyle to (learn to) blend image features while being aware of the object geometry in underwater scenes for waterbody STx. Our proposed architecture (a) integrates a ResNet-based~\cite{koonce2021resnet} encoder-decoder with hierarchical skip connections~\cite{seo2019hierarchical} for effective multi-scale feature fusion; (b) employs a progressive multi-stage training strategy~\cite{wang2018progressive} for stabilized style-content blending; and (c) domain-specific loss functions learn image statistics (color, lightness)~\cite{dong2022underwater}, structural details (SSIM~\cite{hore2010image}, FFT~\cite{brigham1988fast}) and high-level image contents  (VGG~\cite{simonyan2014very}, CLIP~\cite{radford2021learning}). This guided training strategy is leveraged by our proposed DA-WCT mechanism for depth-aware waterbody stylization, ensuring perceptual fidelity and geometric consistency.

\item \textbf{SOTA performance and benchmark evaluation:} We conduct comprehensive experimental evaluations validating that UStyle consistently outperforms leading STx methods such as StyTR-2~\cite{deng2022stytr2}, MCCNet~\cite{wang2020arbitrary}, ASTMAN~\cite{deng2020arbitrary}, IEContraAST~\cite{chen2021artistic}, PhotoWCT2~\cite{chiu2022photowct2}, and StyleID~\cite{chung2024style} across key metrics such as PSNR, SSIM, RMSE, and LPIPS. Ablation studies over all loss components also reveal that our proposed DA-WCT mechanism and domain-specific loss functions can deal with the underwater image distortions to generate perceptually coherent stylization. We also compare the performance with physics-based approaches such as AquaFuse~\cite{siddique2024aquafuse} -- to validate that UStyle-generated images preserve the structural contents of the scene while fusing waterbody styles. These results establish UStyle as a robust and effective solution for no-reference underwater image STx applications.
\end{enumerate}
UStyle provides a robust and efficient solution for no-reference underwater STx, addressing the limitations of traditional artistic and photorealistic STx approaches. Its ability to preserve geometric consistency while achieving realistic waterbody transformations makes it an ideal framework for real-world marine robotics and imaging applications. UStyle opens up significant opportunities for further research in real-time video-based stylization, data augmentation, and interactive augmented and virtual reality (AR/VR) domains.

\vspace{-2.0mm}
\section{Background and Related Work}
\begin{figure*}[t]
    \centering
    \includegraphics[width=0.51\textwidth]{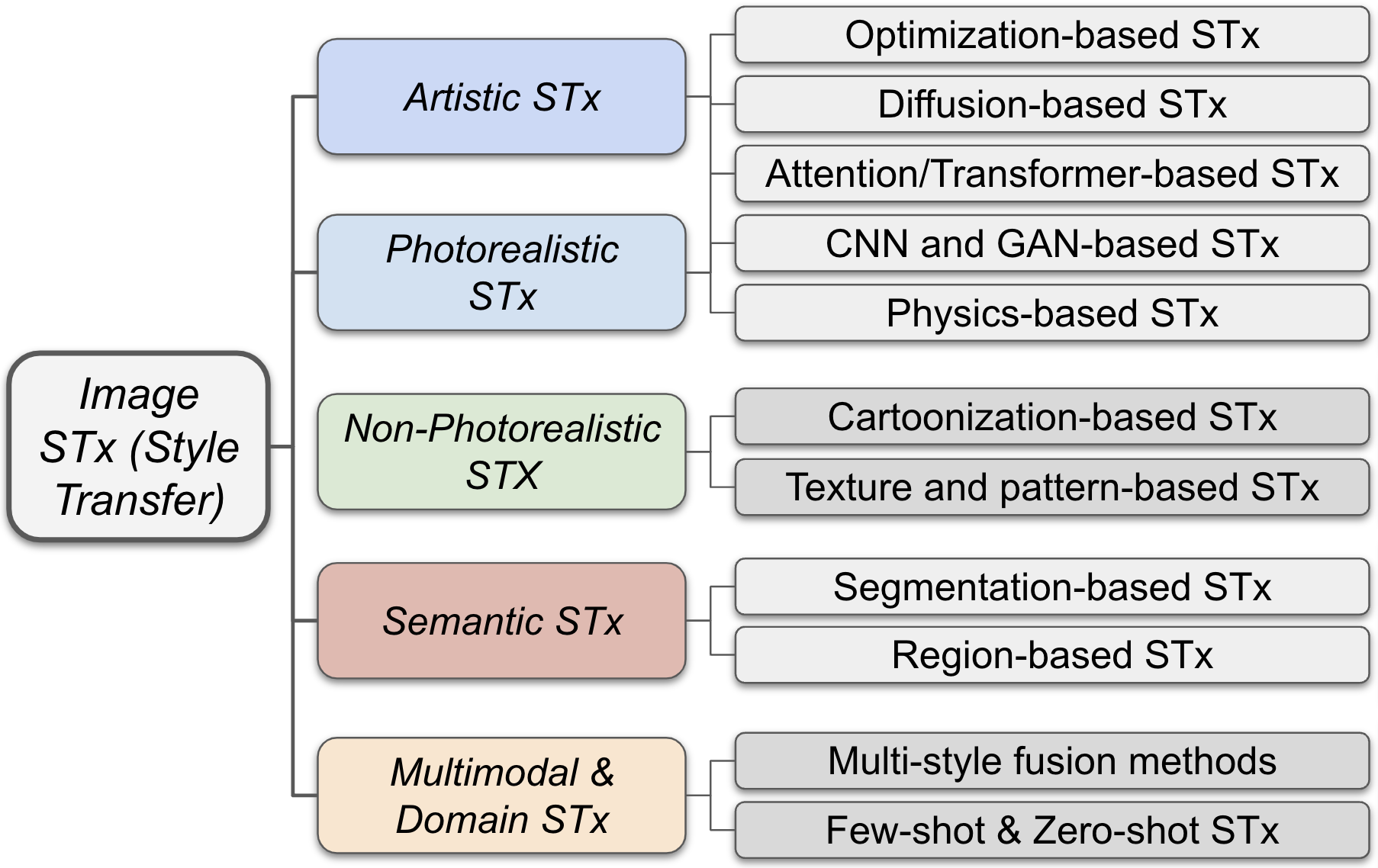}~~
    \footnotesize
    \begin{tabular}[b]{ll}
    \Xhline{2\arrayrulewidth}
    \textbf{SOTA Approaches} & \textbf{Prominent Literature} \\ 
    \Xhline{2\arrayrulewidth}
    \hline
    {Optimization models} & \cite{gatys2015neural,luan2017deep,wang2019research,kim2020deformable, lyu2023infostyler}\\ 
    \hline
    {Diffusion-based models} & \cite{zhang2023inversion,wang2023stylediffusion,chung2024style,li2024diffstyler}\\ 
    \hline
    {Feed-Forward models} & \cite{huang2017arbitrary,li2017universal,li2018closed,chiu2022photowct2,yoo2019photorealistic} \\ 
    \hline
    {Attention models} & \cite{zhang2019self,park2019arbitrary,liu2021adaattn,wang2022esa}\\ 
    \hline
    {Transformers} & \cite{vaswani2017attention,deng2022stytr2,liang2021swinir,wang2022uformer,chen2022domain,han2021transformer, fengxue2023image} \\ 
    \hline
    {CNNs and GANs} & \cite{jiang2021transgan,fu2019dual,lin2020gan,qian2022circular, xu2021learning} \\ 
    \hline
    {Physics-based models} & \cite{banterle2009inverse,ulyanov2018deep,lafortune1997non,horn1981determining,he2010single} \\ 
    \hline
    {Cartoonization models} & \cite{zhao2024gan,ahn2023interactive,chen2024decogan,li2019everyone,wu2021deep}\\ 
    \hline
    {Texture/pattern analysis} & \cite{fu2022language,zheng2024fast,zhang2023style,kim2019transport} \\ 
    \hline
    {Segmentation models} & \cite{lin2021image,zheng2023retinal,ye2023multi,chen2017deeplab,wang2020image,madokoro2022semantic}  \\ 
    \hline
    {Region-based models}  &  \cite{niu2021style,yamaguchi2015region,krutarth2021region,liao2022semantic}\\ 
    \hline
    {Multi-style fusion}  & \cite{ye2023multi,huang2019style,zhang2018multi,yu2024multi} \\ 
    \hline
    {Few-shot \& Zero-shot} &  \cite{fu2023styleadv,huang2020rd,liu2023stylerf}\\
    \hline
    {Vision-language models} &  \cite{kwon2022clipstyler,yang2023zero,cai2023image} \\ 
    \Xhline{2\arrayrulewidth}
    \end{tabular}
\caption{A categorization of the SOTA literature on image-based STx and some prominent methods.}
\label{fig:sota}
\vspace{-2mm}
\end{figure*}

Image style transfer (STx) methods combine the \textit{content} of one image with the \textit{style} of another~\cite{gatys2015neural, liu2024swinit}, enabling applications in computer vision, AR/VR, robotics, and digital art. Existing STx methods include artistic, photorealistic, semantic, and multimodal techniques that employ optimization, feed-forward networks, attention mechanisms, and transformers; see Fig.~\ref{fig:sota}. 

\vspace{-2mm}
\subsection{Artistic and Non-Photorealistic STx} 
Artistic STx blends one image's artistic characteristics, such as brushstrokes, textures, or color palettes, into another image's content. Following the seminal work by Gatys~\etal~\cite{jing2019neural}, many deep visual learning-based networks have been proposed~\cite{huang2017arbitrary,li2017universal,li2018closed}. In particular, Johnson~\etal~\cite{johnson2016perceptual} and Ulyanov~\etal~\cite{ulyanov2016instance} develop feed-forward networks that apply predefined styles in real-time, making the process much faster. One limitation of their approach is that their network needs to be trained for a specific style. Huang and Belongie~\cite{huang2017arbitrary} address this issue with adaptive instance normalization (AdaIN), enabling the transfer of arbitrary styles~\cite{chen2022quality,wang2022clast, chen2022quality}. Li~\etal~\cite{li2017universal} introduce the whitening and coloring transform (WCT), which applies a \textit{whitening} operation to remove content, followed by a \textit{coloring} operation to inject style statistics. A superset of artistic STx is Non-photorealistic rendering (NPR) that focuses on \textit{cartooning}, sketches, and more abstract art~\cite{ahn2023interactive}. Ahn~\etal~\cite{ahn2023interactive} develop an interactive framework that allows users to influence the stylization in real-time, enhancing creativity and personalization~\cite{chen2024decogan}. Additionally, techniques such as language-guided STx~\cite{fu2022language}, and style consistency models~\cite{zhang2023style} expand the possibilities of NPR, making sophisticated stylization accessible to a broader audience~\cite{kim2019transport}.

\vspace{-2mm}
\subsection{Photorealistic STx} 
Photorealistic STx applies stylistic changes while preserving an image's realism and structure, which is essential for scene understanding and robotics. Early methods such as DPST~\cite{luan2017deep} use optimization but struggle with high resolutions. To address this, PhotoWCT~\cite{li2018closed} employs cascaded autoencoders to maintain structure, while PhotoWCT2~\cite{chiu2022photowct2} and wavelet-based WCT2~\cite{yoo2019photorealistic} improve efficiency and detail preservation. Vision transformers also enhance photorealistic STx; for example, StyTr2~\cite{deng2022stytr2} uses dual transformer encoders with content-aware positional encoding (CAPE) for spatial coherence. Unsupervised methods like TransGAN~\cite{jiang2021transgan} and advanced architectures such as TNT~\cite{han2021transformer}, along with diffusion-based models~\cite{zhang2023inversion,li2024diffstyler}, further disentangle style and content for controlled adjustments. Additionally, methods like DIP~\cite{ulyanov2018deep} leverage natural image priors to restore true colors and preserve spatial coherence.

\vspace{-2mm}
\subsection{Semantic and Multimodal STx} 
Semantic STx integrates a high-level understanding of image content to apply styles context-awarely, enhancing the coherence between the transferred style and the underlying content. For instance, Wang~\etal~\cite{wang2020image} use DeepLab~\cite{chen2017deeplab} to preserve object boundaries, while Lin~\etal~\cite{lin2021image} refine this approach with segmentation to maintain structure. Region-based method such as Liao~\etal~\cite{liao2022semantic} enables selective stylization by fusing global semantic context with local features. 
Few-shot and zero-shot approaches, including StyleAdv~\cite{fu2023styleadv}, Huang~\etal~\cite{huang2020rd}, and StyleRF~\cite{liu2023stylerf}, further address the challenge of limited style exemplars. Vision-language models, particularly CLIP~\cite{cai2023image}, integrate textual and visual information to enhance STx. Clipstyler~\cite{kwon2022clipstyler} leverages a patch-wise CLIP loss for text-guided stylization, and Yang~\etal~\cite{yang2023zero} propose a zero-shot contrastive loss for diffusion models~\cite{wang2023stylediffusion} that preserves semantic content without fine-tuning. These advancements significantly improve the flexibility and scalability of STx for real-world applications.

\vspace{-4.9mm}
\subsection{Underwater Domain Style Transfer} Traditionally, \textit{domain STx} methods are used to model underwater image distortions for synthetic data generation~\cite{fabbri2018enhancing,li2017watergan}; the stylized distorted pairs are then used for learning image enhancement~\cite{li2018synthesis,li2019underwater, zhang2023underwater, zhuang2024globally, wu2024underwater, zhou2025net, sun2022underwater, zhou2023underwater, zhang2022underwater, wang2023underwater, hu2023enhancing} or improved object detection~\cite{islam2020fast,yang2022underwater}. These models enhance color correction, clarity, and stylization in a unified learning pipeline by separating content and style across domains. Chen~\etal~\cite{chen2022domain} introduces a domain adaptation framework that leverages transformers for image enhancement~\cite{anil2023novel}. Zhou~\etal~\cite{zhou2023learning} uses a transformer-based model for stylized representation learning of target recognition in underwater acoustic imagery. Our previous work AquaFuse~\cite{siddique2024aquafuse} proposes a physics-guided method for geometry-preserving waterbody fusion for data augmentation, which requires a prior reference image and known waterbody parameters. We address these knowledge gaps by exploring no-reference waterbody STx from comprehensive multimodal databases. 

\vspace{0.0mm}
\begin{figure*}[t]
    \centering
    \begin{subfigure}{0.92\linewidth}
        \centering
       \includegraphics[width=\textwidth]{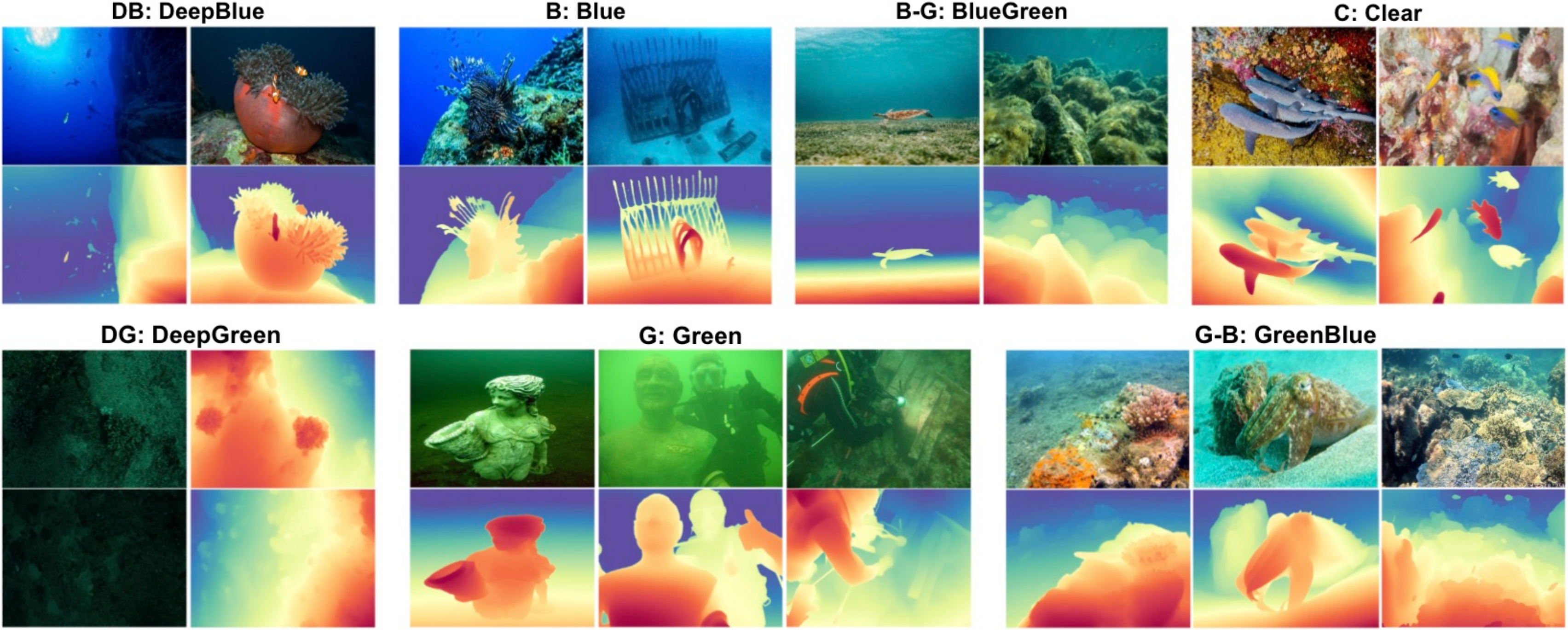}%
    \vspace{-1mm}
    \caption{Sample images and corresponding depth maps for each style category.}
    \label{fig:UF7D_examples}
    \end{subfigure}
    \vspace{1mm}
    
    \begin{subfigure}{0.31\linewidth}
        \centering
       \includegraphics[width=\linewidth]{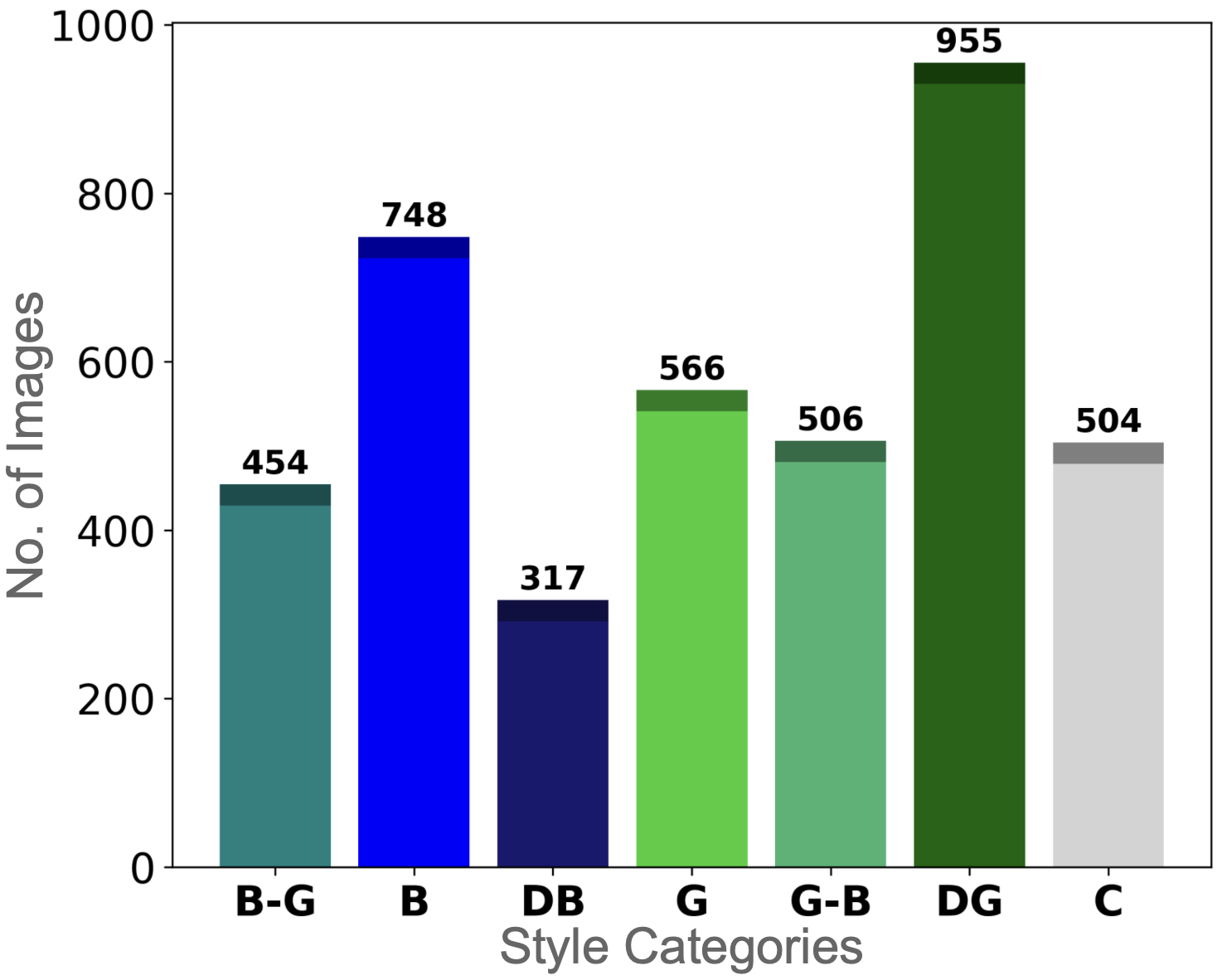}%
      \vspace{-1mm}
      \caption{Distribution of the seven style categories in UF7D.}
      \label{fig:distribution_graph}
    \end{subfigure}
    \hfill
    \begin{subfigure}{0.33\linewidth}
        \centering
        \includegraphics[width=\linewidth]{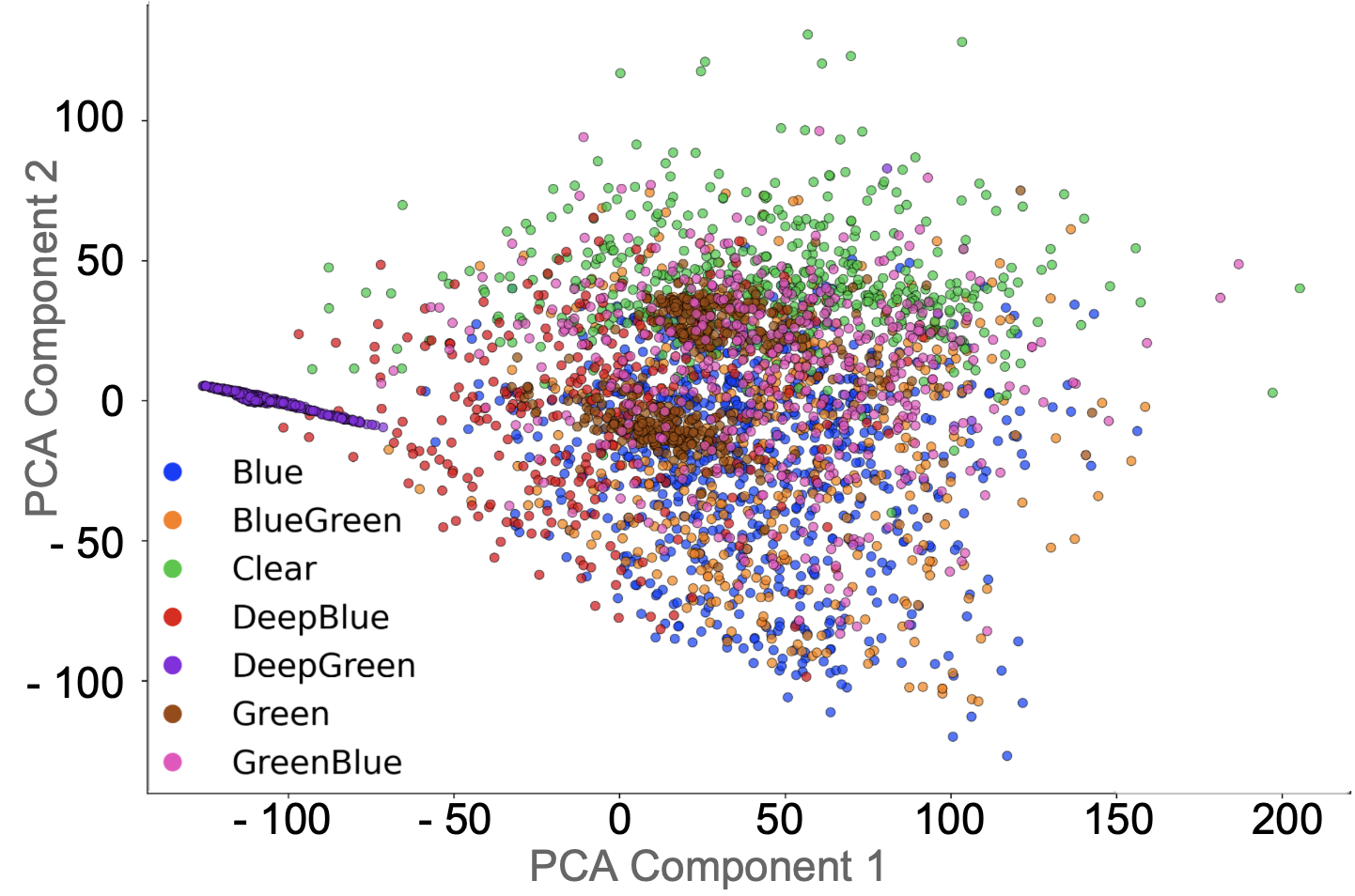}
        \caption{PCA projection of global color characteristics ($92\%$ variance).}
        \label{fig:PCA_scatter_plot}
    \end{subfigure}
    \hfill
    \begin{subfigure}{0.33\linewidth}
        \centering
        \includegraphics[width=\linewidth]{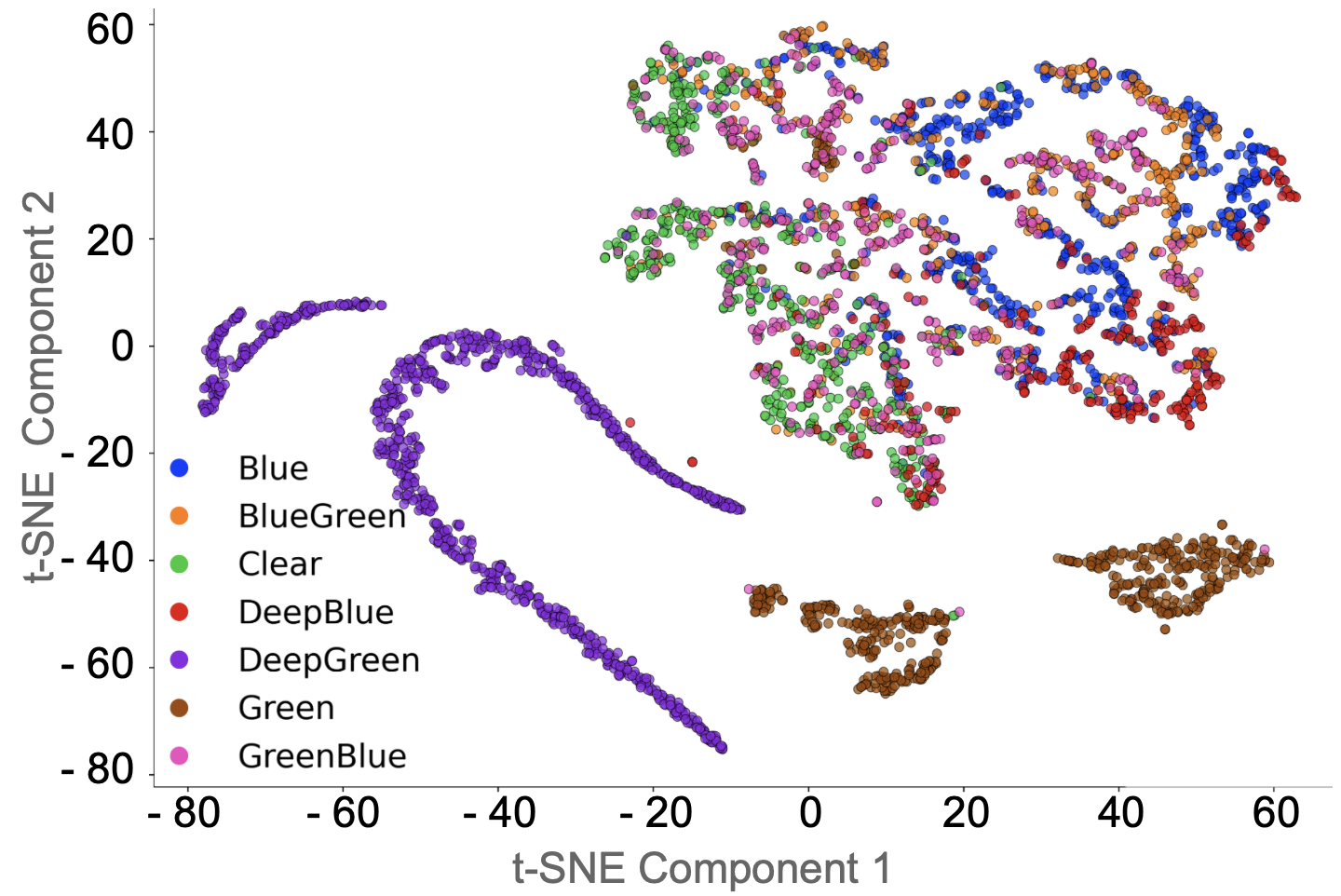}
        \caption{t-SNE visualization of image statistics confirming category distinctions.}
        \label{fig:tSNE_scatter_plot}
    \end{subfigure}
    \vspace{1mm} 
    \caption{Sample images and statistical analyses of the UF7D dataset.}
    \label{fig:UF7D_analysis}
\end{figure*}

\section{UF7D Dataset}

\subsection{Data Preparation and Categorization}\label{sec:uf7d_dataset}
We present \textbf{UF7D}, a high-resolution underwater image dataset comprising seven waterbody styles: Blue (B), Blue-Green (B-G), Clear (C), Deep Blue (DB), Deep Green (DG), Green (G), and Green-Blue (G-B). UF7D combines both benchmark-sourced and newly collected data to ensure broad diversity across underwater environments. Approximately $58\%$ of the images come from existing datasets (UIEB~\cite{li2019underwater}, Sea-Thru~\cite{akkaynak2019sea}, LSUI~\cite{peng2023u}, U45~\cite{li2019underwater}, EUVP~\cite{islam2020fast}), while the other $42\%$ are collected from community and open-access sources (e.g., Unsplash, Flickr), with the focus to introducing diverse scenes and/or waterbody types that are not common in prior benchmarks. The dataset spans varied geographic regions and water types, covering coral reefs, marine species, shipwrecks, pipelines, rocks, and other natural structures. For each image, a relative depth map is generated using DepthAnythingV2~\cite{yang2024depth}. 

UF7D dataset contains $4,050$ images, with over $44\%$ exceeding $1920\times1080$ resolution, enabling fine-grained learning. There are 25 test samples per category, with $625$ content-style evaluation pairs; representative samples and depth maps are shown in Fig.~\ref{fig:UF7D_examples}. To further assess STx generalization, we release a \textbf{UF7D-Challenge} test set with $70$ images across categories, which captures a set of challenging regional, temporal, and sensory variations. The images were collected from online open-access sources and in-field captures, covering diverse locations, lighting, and sensors. This enables testing under unseen environmental and acquisition conditions, following recent evaluation protocols~\cite{Li_Hyunida_2024,boosting_li_2025,LIANG2025104463}.

\vspace{-2mm}
\subsection{Dataset Statistics and Analyses}
UF7D combines benchmark-sourced images with a substantial portion of newly collected, high-resolution data, forming a diverse and well-balanced dataset across different water types and scene compositions. To validate the consistency and distinctness of its seven waterbody categories, we perform global and local statistical analyses as follows:

\begin{enumerate}[label={$\bullet$},nolistsep,leftmargin=*]
\item \textbf{Global color structure}: PCA shows clear separation among the seven waterbody styles, with the first two components explaining $92\%$ of color variance; see Fig.~\ref{fig:PCA_scatter_plot}. The dominant component ($76\%$ variance) separates blue- and green-dominated waters, validating the perceptual categorization.
\item \textbf{Local statistics}: The t-SNE projections shown in Fig.~\ref{fig:tSNE_scatter_plot} reveal distinct, non-overlapping clusters, indicating consistent intra-category characteristics and strong inter-category differences.
\end{enumerate}

These statistics confirm that UF7D is both perceptually and statistically well balanced, with sufficient unseen content to support robust, data-driven learning of underwater \textit{styles}.

\vspace{-3.0mm}
\section{UStyle Model \& Learning Pipeline}\label{sec:model_architecture}
\begin{figure*}[t]
   \centering
   \includegraphics[width=0.94\textwidth]{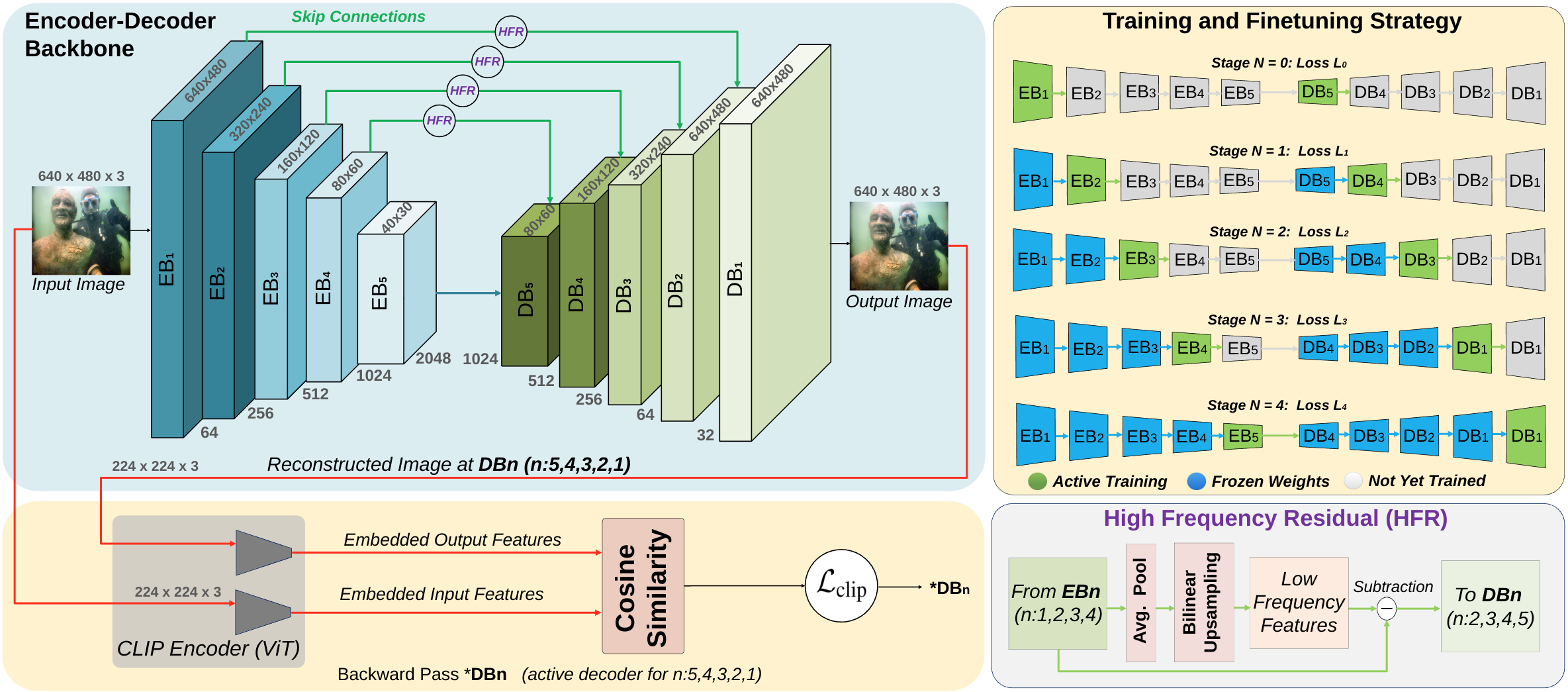}%
   \vspace{-1mm}
    \caption{Overview of our UStyle model, an encoder-decoder with skip connections, high-frequency residual (HFR) paths, and a CLIP-based module aligning features with cues. Multi-stage training progressively trains deeper symmetric blocks to preserve both coarse structure and fine details in waterbody fusion.}
   \label{fig:architecture}
   \vspace{-2mm}
\end{figure*}

We present \textbf{UStyle}, a novel waterbody style transfer framework that integrates a deep encoder-decoder architecture, progressive blockwise training, and a depth-aware stylization module. Given a content image $\mathbf{I}_c \in \mathbb{R}^{H \times W \times 3}$ and a style image $\mathbf{I}_s \in \mathbb{R}^{H \times W \times 3}$, UStyle generates a stylized output $\mathbf{I}_{cs}$ by fusing waterbody characteristics from the style image while preserving scene geometry of the content image. The depth maps $ \mathbf{D}_s $ and $ \mathbf{D}_c $, representing the depth information of the style and content images, respectively, adapt features based on depth. $ \mathbf{D}_s $ guides the extraction of waterbody details from the style image, while $ \mathbf{D}_c $ ensures content-aware stylization by preserving structural consistency and depth-dependent variations during style transfer. The detailed architecture and learning pipeline is outlined in Fig.~\ref{fig:architecture}.

\vspace{-5mm}
\subsection{Encoder–Decoder Backbone} 
We employ a ResNet50~\cite{koonce2021resnet} encoder ($\mathcal{E}$) with five blocks $\mathrm{EB}_{1:5}$ to extract features $\{F_{i,c}\}_{i=0:4}$ and $\{F_{i,s}\}_{i=0:4}$ from the content and style images. These features are passed to a decoder $\mathcal{D}$ via hierarchical skip connections to retain spatial details. Each decoder block $\mathrm{DB}_i$ uses a transposed convolution~\cite{gao2019pixel} for upsampling, followed by convolution, batch normalization, and ReLU activation. Guided by the content depth map $\mathbf{D}_c$, features are fused into ${F}_{i,cs}$—capturing both structural (object geometry) and stylistic (waterbody) elements—via our depth-aware stylization module; see Section~\ref{sec:fusion_process}.

\vspace{-2.0mm}
\subsection{Pre-Training and Domain-specific Finetuning} 
We adopt sequential blockwise training in UStyle over five stages ($N$), freezing incremental $\mathrm{EB}_{i}-\mathrm{DB}_i$ blocks for stability. Initially, the model is pre-trained on a large-scale atmospheric dataset (MSCOCO-2017~\cite{lin2014microsoft}) with a weighted combination of reconstruction loss \cite{rumelhart1986learning}, feature reconstruction loss \cite{johnson2016perceptual}, and perceptual loss. At stage $N$=$0$, only the reconstruction loss is used: $ \mathcal{L}_r = \|\mathbf{I}_c - \hat{\mathbf{I}}_{c}\|_2^2$, where $\hat{\mathbf{I}}_{c}$ is the decoder output. Subsequently, additional feature loss $ \mathcal{L}_{feat} = \|\mathbb{F}_i - \hat{\mathbb{F}}_i\|_2^2$ and perceptual loss $ \mathcal{L}_{percept} = \|\phi(\mathbf{I}_c) - \phi(\hat{\mathbf{I}}_c)\|_2^2$ are added where $\phi(\cdot)$ is a content loss based on VGG-16~\cite{simonyan2014very}.

After pretraining, UStyle is fine-tuned on a large-scale underwater dataset (USOD10K~\cite{hong2023usod10k}) for domain-specific supervision with SSIM loss $\mathcal{L}_{ssim} = 1 - SSIM(\mathbf{I}_c, \hat{\mathbf{I}}_c)$, color consistency loss in LAB image space $\mathcal{L}_{color} = \|\mathrm{LAB}(\mathbf{I}_c) - \mathrm{LAB}(\hat{\mathbf{I}}_c)\|_2^2$, and frequency-domain loss by Fast Fourier Transforms (FFT) $\mathcal{L}_{fft} = \|\mathcal{F}(\mathbf{I}_c) - \mathcal{F}(\hat{\mathbf{I}}_c)\|_2^2$. Moreover, content supervision is incorporated by a CLIP loss~\cite{radford2021learning} to ensure semantic alignment, defined as:    
\begin{equation}
        \mathcal{L}_{clip} = 1 - \mathrm{cosine\_similarity}\Bigl(\mathrm{CLIP}({\mathbf{I}}_c), \mathrm{CLIP}(\hat{\mathbf{I}}_c)\Bigr).
\end{equation}
The overall objective functions are as follows:
\begin{align}    
        \mathcal{L}_0 &= \mathcal{L}_r + \mathcal{L}_{ssim} + \mathcal{L}_{color} + \mathcal{L}_{fft} + \mathcal{L}_{clip} \text{ and} \\
        \mathcal{L}_N &= \mathcal{L}_0 + \mathcal{L}_{feat} + \mathcal{L}_{percept} \text{ for } \quad (N=1,2,3,4).
\end{align}    

\vspace{-2mm}
\subsection{Depth-aware Stylization} \label{sec:fusion_process}
We design a depth-aware stylization function: \textbf{DA-WCT} which extends the multi-scale blending concept from the WCT (whitening and coloring transform) literature~\cite{li2017universal}. DA-WCT integrates a physics-guided waterbody estimation method for depth-guided feature blending via WCT. First, we estimate the \textit{waterbody} from the style image $\mathbf{I}_s$ using a physics-guided method~\cite{siddique2024aquafuse}; from its depth map $\mathbf{D}_s$, we leverage the $5\%$ farthest pixels' median values to estimate the waterbody background ($\mathbf{B}_s$) in $\mathbf{I}_s$. A color filter is then applied to retain only bluish-green pixels, removing red-channel influence due to differential light attenuation underwater. Let $F_c$ and $F_s$ denote the content and style feature maps extracted from the encoder, respectively. We first perform a WCT-based stylization to align the statistics of $F_c$ with those of $F_s$; the stylized feature $\tilde{F}_{cs}$ is computed as $\tilde{F}_{cs} \;=\; \mu_{s} \;+\; C_{s}^{1/2}\,C_{c}^{-1/2}\,\small(F_c - \mu_c\small)$, with $\mu_c$, $\mu_s$ being the channel-wise means and $C_c$, $C_s$ the covariance matrices of $F_c$ and $F_s$, respectively. Let $\mathbf{D}_c$ denote the single-channel content depth map, a global threshold $\tau \in [0,1]$ is computed using Otsu's method \cite{xu2011characteristic}. Next, an adaptive factor $k$ is derived by first smoothing $\mathbf{D}_c$ via average pooling with a $5 \times 5$ kernel, computing the global mean and standard deviation as:
    \begin{align}
    \footnotesize
        \mu_d &= \frac{1}{H\cdot W}\sum_{x} \mathrm{avgpool}\big(\mathbf{D}_{c}(x)\big) \text{ and } \\
        \sigma_d &= \sqrt{\frac{1}{H\cdot W}\sum_{x}\Bigl(\mathrm{avgpool}\big(\mathbf{D}_{c}(x)\big) -\mu_d\Bigr)^2}.
    \label{eq:std_dev}
    \end{align}
With the tunable scaling factors $\tau$ and $k$, the depth weight map for each pixel is computed as: 
\begin{equation}
\begin{split}
        w(x) &= \sigma\Bigl(-k\cdot\bigl(\mathbf{D}_c(x)-\tau\bigr)\Bigr) = \frac{1}{1+\exp\Bigl(k\cdot\bigl(\mathbf{D}_c(x)-\tau\bigr)\Bigr)} \text{.}
\end{split}
\label{eq:sigmoid_func}
\end{equation}

In our formulation, pixels with higher depth values (\ie, farther objects) yield $w(x) \rightarrow 1$ and favor the stylized features $\tilde{F}_{cs}(x)$. In contrast, pixels with lower depth values get $w(x) \rightarrow 0$, preserving more of the original content features $F_c(x)$. When the depth map has low variance and is nearly flat, $\mathbf{D}_c \approx \tau$ which yield $w(x) \rightarrow 0.5$ across the image, so DA-WCT effectively reduces to a standard WCT-style blend. A global style strength parameter $\alpha\in[0,1]$ is applied to control the \textit{degree} of the underlying stylization process as: $F_{cs} \;=\; \alpha\,F_{\text{depth}} \;+\; \bigl(1-\alpha\bigr)\,F_c$.

Furthermore, to capture both global and local style characteristics, our DA-WCT function is applied at multiple scales $s\in\{1.0,\,0.75,\,0.5,\,0.25\}$, with the fused features at each scale, to obtain multi-scale features $F_{\text{multi}}$. At each scale (stage) of the decoder, these \textit{fused features} replace the corresponding \textit{content features} with multi-scale fusion. The stylized image is then constructed as $\mathcal{D}$ as $\mathbf{I}_{cs} = \mathcal{D}\big(F_{\text{multi}}\big)$. Finally, a guided filter named GIF smoothing~\cite{he2012guided} is applied as a post-processing step to enhance the visual consistency of the final output image: $\mathbf{I}_{stylized} = \text{GIF}(\mathbf{I}_{cs}) = \text{GIF}\big(\mathcal{D}(F_{\text{multi}})\big)$.

\begin{figure*}[t]
    \centering
    \includegraphics[width=0.96\textwidth]{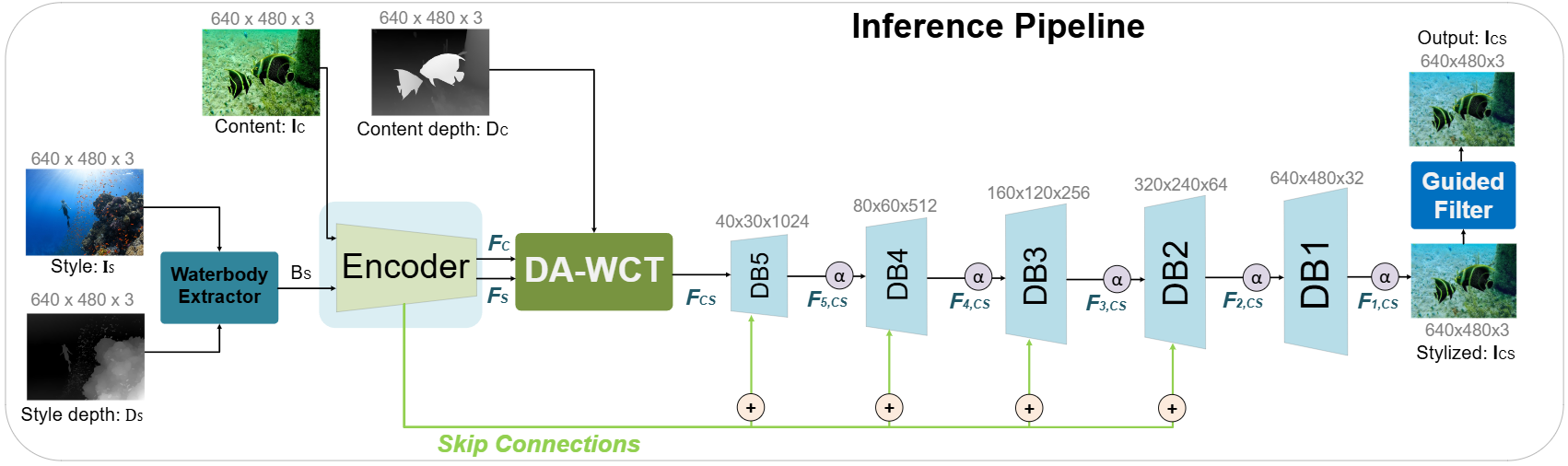}%
    \caption{Inference pipeline of UStyle: The content and style images and their depth maps are first processed by a ResNet-based encoder to extract multi-scale features. These are then aligned using our DA-WCT and progressively decoded via a multi-stage decoder and smoothed by a guided filter.
    }
    \label{fig:inference}
    \vspace{-2mm}
\end{figure*}

\vspace{1mm}
\noindent
\textbf{Implementation Details.} UStyle is implemented in PyTorch with OpenCV and Torch libraries. We use Adam optimizer with an initial learning rate of $10^{-4}$ for supervised learning. Training is conducted on a Linux system with $4$\,GPUs and $32$\,GB of RAM. At the first stage of training, the encoder-decoder network is pre-trained on the MSCOCO-2017 with $118$\,K training samples for $10$ epochs. It is then fine-tuned on USOD10K underwater dataset with $10$\,K examples for another $10$ epochs. The whole training and fine-tuning takes $82$ hours to complete. Once trained, the frozen graph is used for inference on a single GPU with $16$\,GB of RAM; the inference pipeline is shown in Fig.~\ref{fig:inference}.%

\vspace{-2mm}
\section{Experimental Results}\label{subsec:experiment}
\vspace{-2mm}
\subsection{Baseline Models and Metrics}
Seven state-of-the-art (SOTA) models from artistic and photorealistic STx domains are used for comparison, including learning-based models: StyTR-2~\cite{deng2021stytr2}, ASTMAN~\cite{deng2020arbitrary}, MCCNet~\cite{wang2020arbitrary}, IEContraAST~\cite{qi2020iecontrasast}, PhotoWCT2~\cite{chiu2022photowct2}, StyleID~\cite{chung2024style}, and the underwater physics-based method: AquaFuse~\cite{siddique2024aquafuse}. All models are fine-tuned on UF7D for $160$\,K iterations with a batch size of $8$, using their default hyperparameters for fairness.
For evaluation, we define four cross-domain underwater style transfer test cases: Blue (B) $\rightarrow$ Green (G), Green (G) $\rightarrow$ Blue (B), Deep Blue (DB) $\rightarrow$ Deep Green (DG), and Deep Green (DG) $\rightarrow$ Deep Blue (DB). Each case consists of $25$ content images and $25$ style images, yielding $25 \times 25 = 625$ unique test scenes per case, for a total of $4 \times 625 = 2,500$ test scenes.

\noindent\textbf{Evaluation Metrics.}
We evaluate UStyle and all baseline methods using five standard image-quality metrics. These are: (\textbf{i}) RMSE~\cite{chai2014root} for pixel-level deviation; (\textbf{ii}) PSNR~\cite{hore2010image} for reconstruction fidelity; (\textbf{iii}) SSIM~\cite{wang2004image} to evaluate luminance, contrast, and structural consistency; (\textbf{iv}) GMSD~\cite{xue2014gradient} for edge preservation and local gradient similarity; and (\textbf{v}) LPIPS~\cite{zhang2018unreasonable} to capture high-level perceptual similarity using deep feature embeddings. Together, these metrics provide a comprehensive assessment of both structural accuracy and perceptual quality, with all results computed on images normalized to $640\times480$ using official metric implementations.

\begin{figure*}[t]
   \centering
   \includegraphics[width=0.98\textwidth]{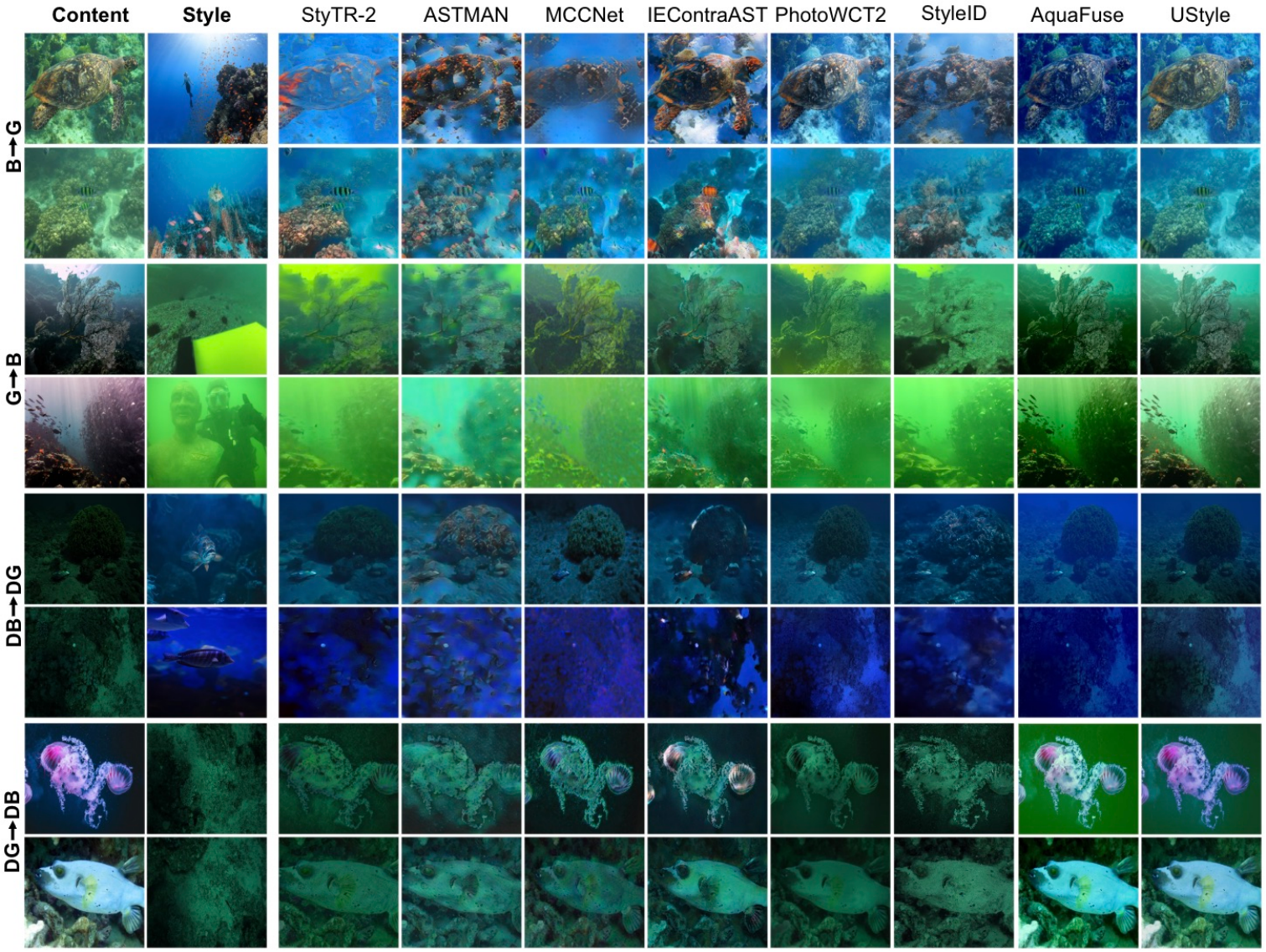}%
   \vspace{-1mm}
   \caption{Qualitative comparison of underwater image style transfer (STx) results across different methods. The first two columns show content and style images, followed by stylized outputs from StyTR-2~\cite{deng2022stytr2}, ASTMAN~\cite{deng2020arbitrary}, MCCNet~\cite{deng2021arbitrary}, IEContraAST~\cite{chen2021artistic}, PhotoWCT2~\cite{chiu2022photowct2}, StyleID~\cite{chung2024style}, AquaFuse~\cite{siddique2024aquafuse}, and UStyle. Each row shows two examples of: B $\rightarrow$ G, G $\rightarrow$ B, DB $\rightarrow$ DG, and DG $\rightarrow$ DB transformations.
   }  
   \label{fig:qualitative}
\end{figure*}

\vspace{-2mm}
\subsection{Qualitative Results}
Fig.~\ref{fig:qualitative} presents qualitative comparisons across diverse underwater scenes, evaluating four primary waterbody STx cases: B$\rightarrow$G, G$\rightarrow$B, DB$\rightarrow$DG, and DG$\rightarrow$DB. Existing artistic STx methods (StyTR-2, ASTMAN, MCCNet, StyleID and IEContraAST) struggle to preserve underwater scene realism. Notably, StyTR-2, StyleID, and ASTMAN introduce noticeable artifacts and texture inconsistencies, distorting natural structures. While MCCNet and IEContraAST improve content preservation, they tend to oversaturate stylized regions, introducing subtle color artifacts. The photorealistic STx method PhotoWCT2 maintains structural integrity in most cases; however, it fails to fully capture deep-water transformations (\eg, DB$\leftrightarrow$DG). Meanwhile, the physics-based AquaFuse ensures better underwater color consistency, although it requires a carefully curated reference image for optimal performance.

The proposed UStyle method addresses these limitations by incorporating depth awareness into the STx process. Unlike conventional approaches that apply stylization without spatial understanding, UStyle leverages depth supervision to ensure geometric consistency. This advantage is particularly evident in challenging deep-water transitions (DB$\leftrightarrow$DG), where other methods struggle to balance structural and color fidelity. We hypothesize that UStyle's superior performance stems from its adaptive multi-stage training and domain-aware waterbody stylization mechanisms (see Sec.~\ref{ablation} for ablation results). Overall, UStyle produces visually coherent, artifact-free stylized outputs that adapt to waterbody style distributions while preserving object geometry and structural integrity.

\subsection{Quantitative Results}
We conduct a comprehensive quantitative evaluation using five complementary metrics across various underwater STx scenarios from the UF7D test set. As presented in Table~\ref{tab:rmse_psnr_ssim}(a), UStyle achieves $4\times$ to $6\times$ lower RMSE rates across all scenarios compared to SOTA artistic and photorealistic STx methods. It also consistently outperforms competing approaches in PSNR and SSIM, as shown in Table~\ref{tab:rmse_psnr_ssim}(b). The closest competitor, AquaFuse, requires a carefully selected reference image and prior information such as incident angle and attenuation characteristics. In contrast, UStyle achieves improvements of $15\%$ (RMSE), $5.29\%$ (PSNR), and $5.71\%$ (SSIM) over AquaFuse while operating without any reference or prior knowledge — demonstrating its robustness and accuracy for waterbody STx.

We further provide a quantitative comparison of perceptual metrics, including GMSD and LPIPS, as summarized in Table~\ref{tab:gmsd_lpips}. The results demonstrate that UStyle achieves superior preservation of both low-level image statistics and high-level perceptual fidelity. Notably, it attains the lowest LPIPS scores, with values of $0.1926$ (B$\rightarrow$G), $0.2235$ (G$\rightarrow$B), $0.1342$ (DB$\rightarrow$DG), and $0.1373$ (DG$\rightarrow$DB). In comparison, PhotoWCT2 exhibits an LPIPS range of $0.1676$–$0.2618$, while artistic stylization methods such as ASTMAN and MCCNet perform less effectively, exceeding 0.3 in most transformations. Furthermore, UStyle consistently achieves the lowest GMSD scores across all test cases. The accompanying heatmap visualizations in Table~\ref{tab:gmsd_lpips}(c) further illustrate UStyle’s superior performance for combined deviations of each model based on GMSD and LPIPS. These findings align with our qualitative analyses, reinforcing that UStyle enables robust and perceptually accurate waterbody style transfer while preserving the structural integrity of the scene.

\begin{table*}[t]
    \centering
    \caption{Quantitative comparison of averaged RMSE ($\downarrow$), PSNR ($\uparrow$), and SSIM ($\uparrow$) across different underwater STx methods are conducted on $2,500$ test samples in UF7D. The best score is highlighted in {\color{blue} \textbf{bold blue}}, while the second-best is shown in {\color{blue} blue}. See Table~\ref{tab:gmsd_lpips} for GMSD and LPIPs evaluation.}
    \label{tab:rmse_psnr_ssim}   
    \vspace{-4mm}
    \begin{minipage}{0.50\textwidth}
        \centering
        \footnotesize
        \begin{tabular}{l|cccc}
        \toprule
        \cellcolor{gray!20}{(\textbf{a}) RMSE $\downarrow$} & \textbf{B $\rightarrow$ G} & \textbf{G $\rightarrow$ B} & \textbf{DB $\rightarrow$ DG} & \textbf{DG $\rightarrow$ DB} \\
        \midrule
        ASTMAN~\cite{deng2020arbitrary}         & $52.29$      & $54.53$      & $48.91$     & $63.74$      \\
        
        IEContraAST~\cite{qi2020iecontrasast}     & $47.32$      & $50.54$      & $58.17$      & $69.49$      \\
        MCCNet~\cite{wang2020arbitrary}         & $44.64$      & $53.12$      & $46.04$      & $60.71$      \\
        PhotoWCT2~\cite{chiu2022photowct2}       & $34.68$      &$34.80$     & $31.89$      & $26.66$     \\
        StyTR-2~\cite{deng2021stytr2}         & $37.08$      & $40.61$      & $47.18$      & $54.01$      \\
        {StyleID~\cite{chung2024style}} & 48.44 & 53.90 & 55.05 & 52.55  \\
    AquaFuse~\cite{siddique2024aquafuse}          & \textcolor{blue}{$10.45$}      & \textcolor{blue}{$10.19$}     & \textcolor{blue}{$10.15$}      & \textcolor{blue}{$10.24$}      \\
        UStyle (ours)     & \textbf{\textcolor{blue}{9.39}}      & \textbf{\textcolor{blue}{9.35}}     & \textbf{\textcolor{blue}{9.50}}      & \textbf{\textcolor{blue}{8.96}}      \\
        \bottomrule
        \end{tabular}
    \end{minipage}%
    \hfill
    \begin{minipage}{0.48\textwidth}
    \centering
    \vspace{4mm}
    \includegraphics[width=0.9\textwidth]{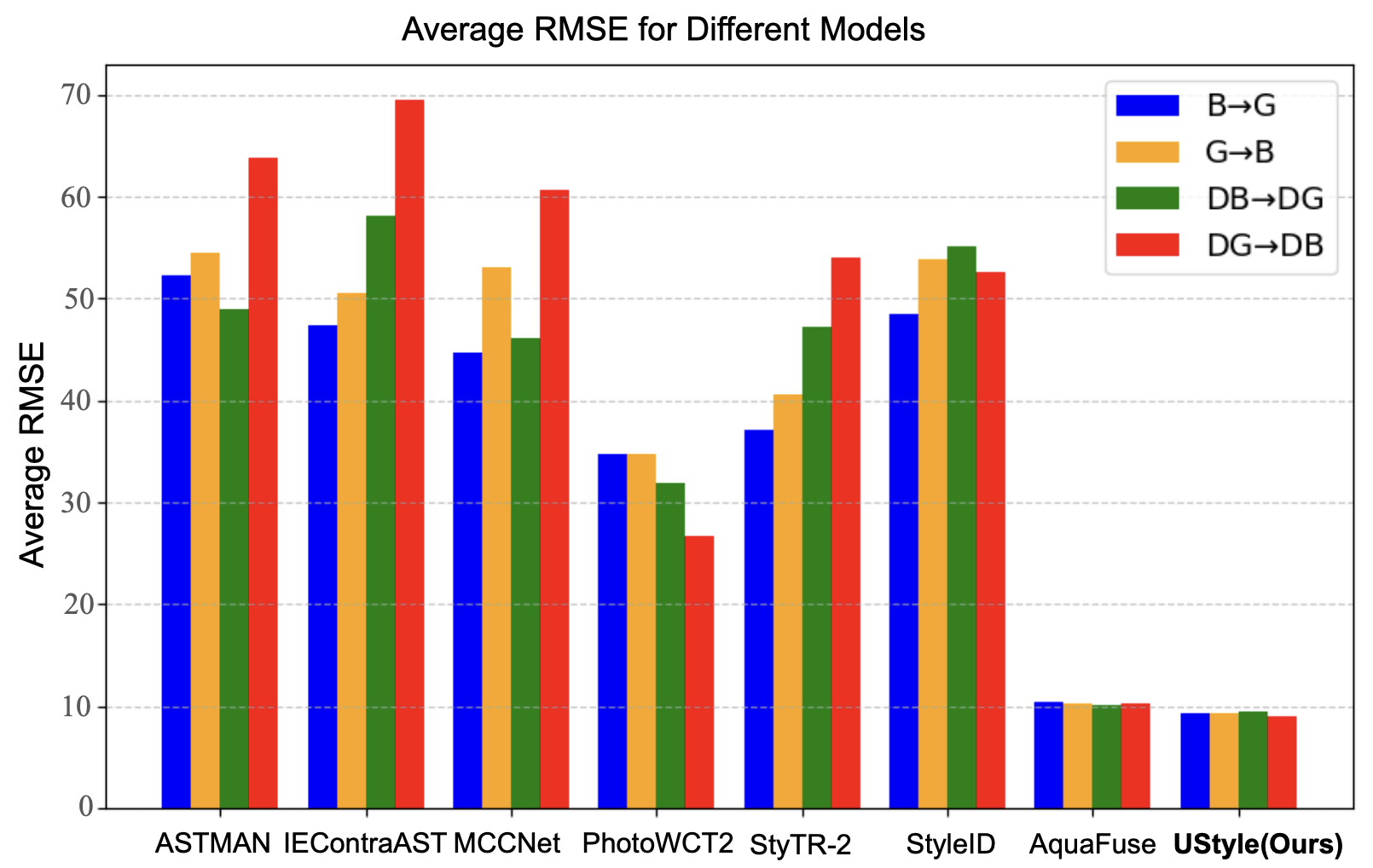}
    \end{minipage}

    \begin{minipage}{0.52\textwidth} 
        \centering
        \footnotesize 
        \renewcommand{\arraystretch}{1.1} 
        \begin{tabular}{l|cccc}
        \toprule
        \cellcolor{gray!20}{(\textbf{b}) PSNR $\uparrow$} & \textbf{B $\rightarrow$ G} & \textbf{G $\rightarrow$ B} & \textbf{DB $\rightarrow$ DG} & \textbf{DG $\rightarrow$ DB} \\
        \midrule
        ASTMAN~\cite{deng2020arbitrary}         & $14.95$      & $14.37$      & $14.99$      & $13.16$     \\
        IEContraAST~\cite{qi2020iecontrasast}   & $15.90$      & $14.40$      & $13.82$      & $12.10$      \\
        MCCNet~\cite{wang2020arbitrary}        & $15.81$      & $14.01$      & $15.63$      & $13.35$      \\
        PhotoWCT2~\cite{chiu2022photowct2}      & $19.75$      & $18.70$      & $18.95$      & $20.05$      \\
        StyTR-2~\cite{deng2021stytr2}          & $17.86$      & $16.67$      & $15.59$      & $14.58$      \\
        {StyleID~\cite{chung2024style}} & 14.72 & 13.73& 14.32 & 14.62 \\
        AquaFuse~\cite{siddique2024aquafuse}   & \textcolor{blue}{$27.77$}      & \textcolor{blue}{$28.00$}     & \textcolor{blue}{$28.03$}      & \textcolor{blue}{$27.94$}      \\
        UStyle (ours)                                & \textbf{\textcolor{blue}{28.73}} & \textbf{\textcolor{blue}{28.72}} & \textbf{\textcolor{blue}{28.64}} & \textbf{\textcolor{blue}{29.14}} \\
        \bottomrule
        \end{tabular}

        \vspace{1em} 
        \footnotesize 
        \begin{tabular}{l|cccc}
        \toprule
        \cellcolor{gray!20}{(\textbf{c}) SSIM $\uparrow$} & \textbf{B $\rightarrow$ G} & \textbf{G $\rightarrow$ B} & \textbf{DB $\rightarrow$ DG} & \textbf{DG $\rightarrow$ DB} \\
        \midrule
        ASTMAN~\cite{deng2020arbitrary}       & $0.8026$      & $0.6334$      & $0.8150$      & $0.6397$      \\
        IEContraAST~\cite{qi2020iecontrasast} & $0.7641$      & $0.6304$      & $0.7887$      & $0.6033$      \\
        MCCNet~\cite{wang2020arbitrary}       & $0.8095$      & $0.6302$      & $0.8290$      & $0.6570$      \\
        PhotoWCT2~\cite{chiu2022photowct2}    & $0.8675$      & $0.7819$      & \textcolor{blue}{$0.8324$} & $0.8030$  \\
        StyTR-2~\cite{deng2021stytr2}        & $0.8494$      & $0.6937$      & $0.8199 $     & $0.7125$      \\
        {StyleID~\cite{chung2024style}} & 0.3528 & 0.3503 & 0.2800 & 0.3527 \\
        AquaFuse~\cite{siddique2024aquafuse} & \textcolor{blue}{$0.9337$}      & \textcolor{blue}{$0.8708$}      & $0.7394$      & \textcolor{blue}{$0.8812$}      \\
        UStyle (ours)                               & \textbf{\textcolor{blue}{0.9613}} & \textbf{\textcolor{blue}{0.8818}} & \textbf{\textcolor{blue}{0.8389}} & \textbf{\textcolor{blue}{0.9126}} \\
        \bottomrule
        \end{tabular}
    \end{minipage}%
    \hfill 
    \begin{minipage}{0.45\textwidth} 
        \centering
        \vspace{1em} 
        \includegraphics[width=\textwidth]{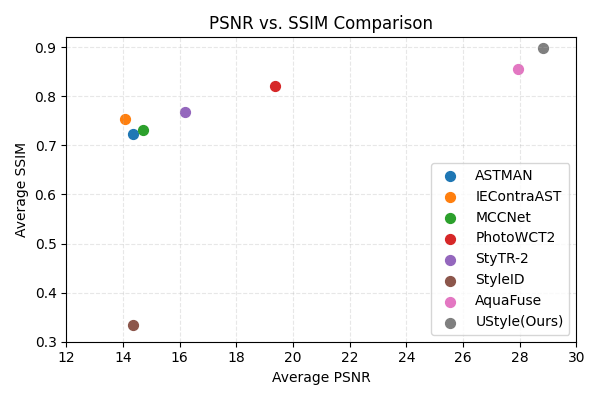} 
        \subcaption*{(\textbf{d}) Average RMSE scores are shown on (top barchart) and the PSNR$-$SSIM scores are visualized in a 2D grid (bottom); evident from these results, UStyle achieves better overall performance than existing STx methods.}
    \end{minipage}
\end{table*}

\begin{table*}[t]
    \centering
    \caption{{Quantitative comparison of GMSD ($\downarrow$) and LPIPS ($\downarrow$) across different underwater STx methods, averaged over $2,500$ test samples in UF7D. }}
    \label{tab:gmsd_lpips}
    \vspace{-2mm}
    \begin{minipage}{0.52\textwidth} 
        \centering
        \footnotesize 
        \renewcommand{\arraystretch}{1.1} 
        \begin{tabular}{l|cccc}
        \toprule
        \cellcolor{gray!20}{(\textbf{a}) GMSD $\downarrow$} & \textbf{B $\rightarrow$ G} & \textbf{G $\rightarrow$ B} & \textbf{DB $\rightarrow$ DG} & \textbf{DG $\rightarrow$ DB} \\
        \midrule
        ASTMAN~\cite{deng2020arbitrary}         & $0.0693$      & $0.0318$      & $0.0508$      & \textbf{\textcolor{blue}{0.0179}} \\
        IEContraAST~\cite{qi2020iecontrasast}   & $0.0694$      & $0.0300$      & $0.0556$      & \textcolor{blue}{$0.0187$}      \\
        MCCNet~\cite{wang2020arbitrary}         & $0.0262$      & \textbf{\textcolor{blue}{0.0107}} & $0.0386$ & $0.0386$      \\
        PhotoWCT2~\cite{chiu2022photowct2}      & $0.0622$      & $0.0725$      & $0.0762$ & $0.0386$ \\
        StyTR-2~\cite{deng2021stytr2}          & $0.0551$      & \textcolor{blue}{$0.0287$} & $0.0380$ & $0.0187$ \\
        {StyleID~\cite{chung2024style}} & 0.0710 & 0.0321 & 0.0685 & 0.0403  \\
        AquaFuse~\cite{siddique2024aquafuse}   & \textcolor{blue}{$0.0252$} & $0.0631$     & \textcolor{blue}{$0.0201$} & $0.0471$      \\
        UStyle (ours)                                & \textbf{\textcolor{blue}{0.0227}} & $0.0564$ & \textbf{\textcolor{blue}{0.0200}} & $0.0431$ \\
        \bottomrule
        \end{tabular}

        \vspace{1em} 
        \footnotesize 
        \begin{tabular}{l|cccc}
        \toprule
        \cellcolor{gray!20}{(\textbf{b}) LPIPS $\downarrow$} & \textbf{B $\rightarrow$ G} & \textbf{G $\rightarrow$ B} & \textbf{DB $\rightarrow$ DG} & \textbf{DG $\rightarrow$ DB} \\
        \midrule
        ASTMAN~\cite{deng2020arbitrary}       & $0.2992$      & $0.4849$      & $0.3769$      & $0.4091$      \\
        IEContraAST~\cite{qi2020iecontrasast} & $0.2762$      & $0.4835$      & $0.3896$      & $0.4518$      \\
        MCCNet~\cite{wang2020arbitrary}       & $0.2838$      & $0.4976$      & $0.3677$      & $0.3989$      \\
        PhotoWCT2~\cite{chiu2022photowct2}    & \textbf{\textcolor{blue}{0.1676}}  & \textcolor{blue}{$0.2618$} & \textcolor{blue}{$0.2409$} & \textcolor{blue}{$0.1755$} \\
        StyTR-2~\cite{deng2021stytr2}        & $0.2372$      & $0.3612$      & $0.3624$      & $0.3261$      \\
        {StyleID~\cite{chung2024style}} & 0.5164 & 0.4091 & 0.3710 & 0.4618 \\
        AquaFuse~\cite{siddique2024aquafuse}  & $0.4274$      & $0.3902$     & $0.5597$      & $0.3789$     \\
        UStyle (ours)                               & \textcolor{blue}{$0.1926$} & \textbf{\textcolor{blue}{0.2235}} & \textbf{\textcolor{blue}{0.1342}} & \textbf{\textcolor{blue}{0.1373}} \\
        \bottomrule
        \end{tabular}
    \end{minipage}%
    \hfill 
    \begin{minipage}{0.45\textwidth} 
        \centering
        \vspace{-1em} 
        \includegraphics[width=0.9\textwidth]{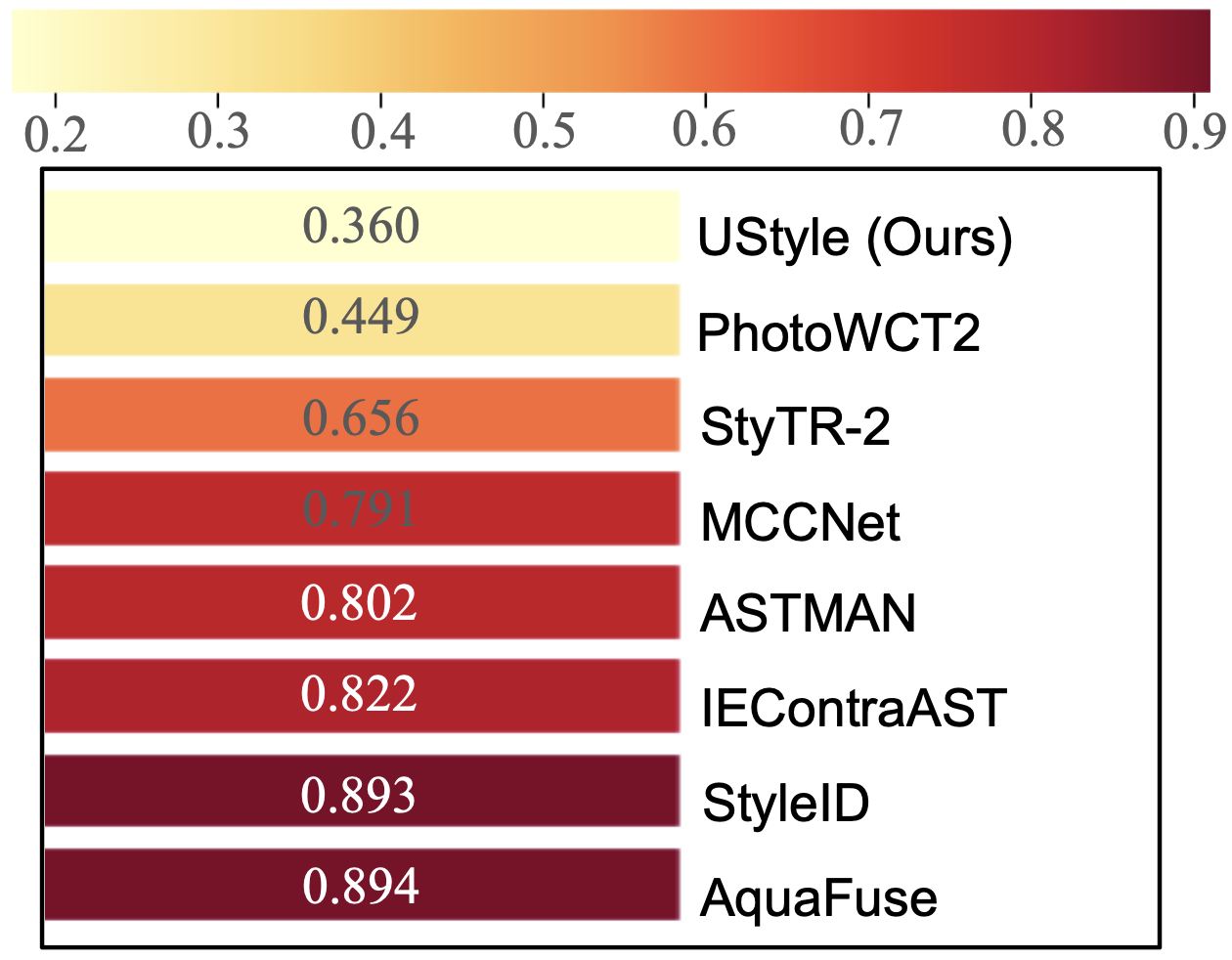}%
        \vspace{2mm}
        \subcaption*{(\textbf{c}) A heatmap illustrating the combined deviation, which represents the overall distortion by each model, computed by GMSD and LPIPS scores. Here, lower values indicate better overall performance; UStyle achieves the lowest score, $19.8\%$ lower than the closest competitor.}
    \end{minipage}
\end{table*}

\begin{figure}[t]
    \centering
    \includegraphics[width=0.75\textwidth]{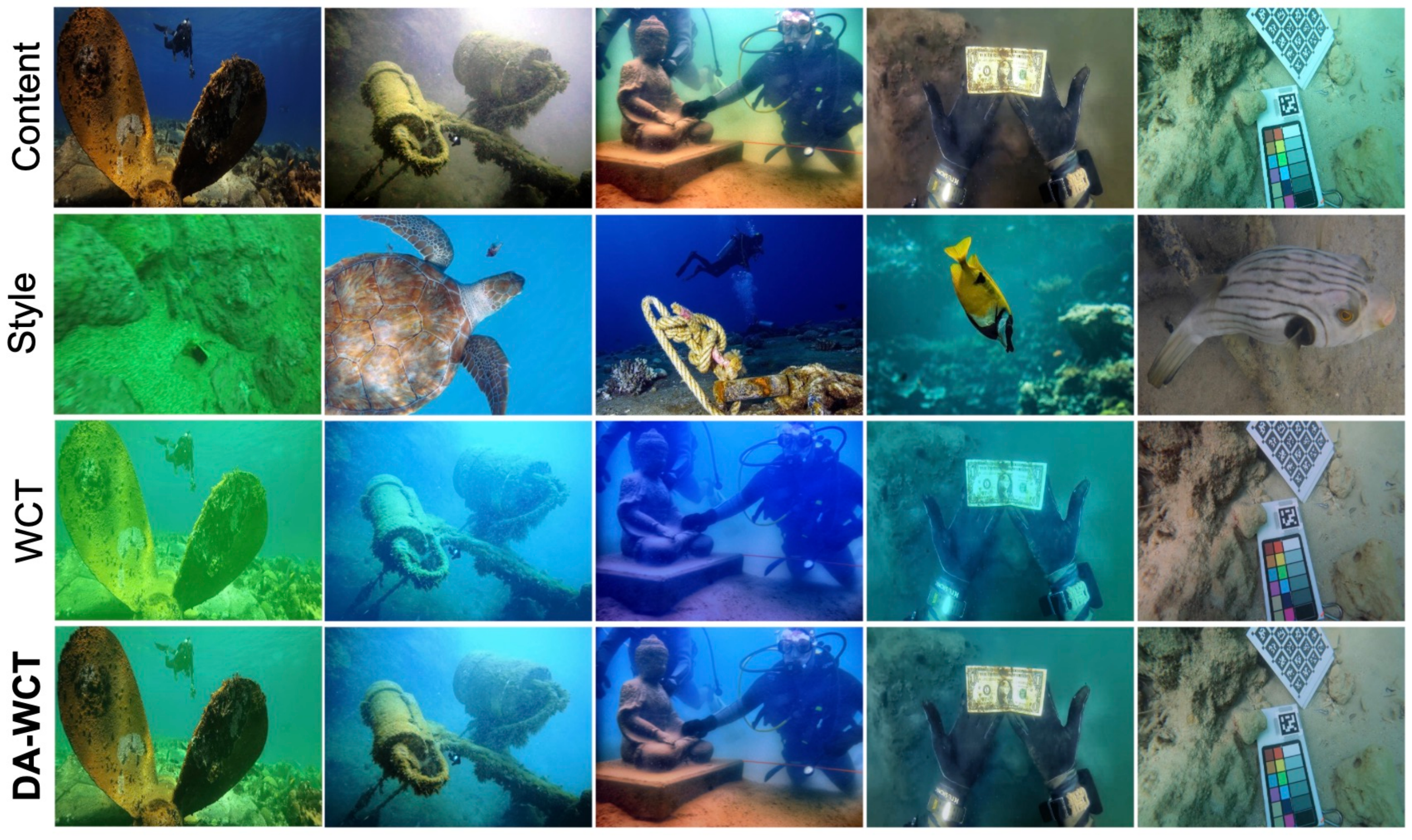}%
    \vspace{-1mm}
    \caption{A few ablation results illustrating the impact of depth-guidance in the stylization process. With the proposed DA-WCT blending, stylized outputs exhibit consistent blending, both on foreground objects and background waterbody regions. In contrast, we observe a global averaging effect while using only WCT and over-saturated pixels for deep-water (DB/DG) styles.}
    \label{fig:gradcam}
\end{figure}

\begin{table*}[ht]
\centering
\caption{{For ablation on each loss term's contributions, we compare the performance of UStyle without: $\mathcal{L}_{ssim}$, $ \mathcal{L}_{color}$, $ \mathcal{L}_{fft}$, $ \mathcal{L}_{percept}$, $ \mathcal{L}_{clip}$ and $ {L}_{3}$, compared to the full model. Each of these variants is evaluated with WCT/DA-WCT on PSNR (\textuparrow), SSIM (\textuparrow), and RMSE (\textdownarrow) metrics.}}
\label{tab:ablation_table}
\renewcommand{\arraystretch}{1.2}
\footnotesize
\begin{tabular}{c|l|cc|cc|cc|cc}
\toprule
\multicolumn{2}{c|}{\cellcolor{gray!10}{\textbf{Data}}} & \multicolumn{2}{c|}{\cellcolor{gray!10}{B $\rightarrow$ G}} & \multicolumn{2}{c}{\cellcolor{gray!10}{G $\rightarrow$ B}} & \multicolumn{2}{c|}{\cellcolor{gray!10}{DB $\rightarrow$ G}} & \multicolumn{2}{c}{\cellcolor{gray!10}{DG $\rightarrow$ B}} \\
\midrule
\textbf{Metric} & \textbf{Model} & WCT & DA-WCT & WCT & DA-WCT & WCT & DA-WCT & WCT & DA-WCT \\
\hline

\multirow{3}{*}{{PSNR}}
& 1: UStyle (w/o $\mathcal{L}_{SSIM}$) & $27.48$  & $28.47$  & $27.84$  & $28.65$  & $27.37$  & $27.73$  & $27.85$  & $28.70$  \\
 & 2: UStyle (w/o $\mathcal{L}_{color}$ ) & $27.58$  & $28.62$  & $27.96$  & $28.74$  & $27.73$  & $28.22$  & $27.92$  & $28.76$  \\
 & 3: UStyle (w/o $\mathcal{L}_{fft}$ )     & $27.43$  & $28.56$  & $27.56$  & $28.15$  & $27.29$  & $27.62$  & $28.02$  & $28.84$  \\
 & 4: UStyle (w/o $\mathcal{L}_{percept}$) & $27.93$  & $28.76$  & $28.29$  & $28.80$  & $27.49$  & $27.93$  & $28.15$  & $28.95$  \\
 & 5: UStyle (w/o $\mathcal{L}_{clip}$) & $28.03$  & $28.69$  & $28.24$  & $28.69$  & $27.85$  & $28.00$  & $28.39$  & $29.09$  \\
 & 6: UStyle (w/o $L3$)   & $27.96$  & $28.36$  & $28.21$  & $28.64$  & $27.84$  & $27.91$  & $28.43$  & $28.97$  \\
 & 7: \textbf{UStyle (Full)}     & $28.09$  & $28.73$  & $28.24$  & $28.72$  & $27.91$  & $28.64$  & $28.43$  & $29.14$  \\
\hline

\multirow{3}{*}{{SSIM}}
&1: UStyle (w/o $\mathcal{L}_{SSIM}$) & $0.9285$  & $0.9434$  & $0.7975$  & $0.8693$  & $0.5990$  & $0.7771$  & $0.8402$  & $0.8981$  \\
 & 2: UStyle (w/o $\mathcal{L}_{color}$)  & $0.9307$  & $0.9594$  & $0.8001$  & $0.8770$  & $0.6157$  & $0.7961$  & $0.8370$  & $0.9002$  \\
 & 3: UStyle (w/o $\mathcal{L}_{fft}$ )     & $0.9291$  & $0.9516$  & $0.7990$  & $0.8688$  & $0.6014$  & $0.7912$  & $0.8391$  & $0.9012$  \\
 & 4: UStyle (w/o $\mathcal{L}_{percept}$) & $0.9309$  & $0.9540$  & $0.8102$  & $0.8775$  & $0.6216$  & $0.8059$  & $0.8456$  & $0.9080$  \\
 & 5: UStyle (w/o $\mathcal{L}_{clip}$) & $0.9319$  & $0.9549$  & $0.8090$  & $0.8758$  & $0.6365$  & $0.8046$  & $0.8421$  & $0.9102$  \\
 & 6: UStyle (w/o $L3$)   & $0.9337$  & $0.9590$  & $0.8004$  & $0.8675$  & $0.6087$  & $0.7780$  & $0.8388$  & $0.9069$  \\
 & 7: \textbf{UStyle (Full)}     & $0.9349$  & $0.9613$  & $0.8260$  & $0.8818$  & $0.6727$  & $0.8389$  & $0.8506$  & $0.9126$  \\
\hline

\multirow{3}{*}{{RMSE}}
&1: UStyle (w/o $\mathcal{L}_{SSIM}$) & $10.77$  & $9.61$  & $10.33$  & $9.42$  & $10.91$  & $10.47$  & $10.32$  & $9.36$  \\
 & 2: UStyle (w/o $\mathcal{L}_{color}$)  & $10.65$  & $9.45$  & $10.19$  & $9.32$  & $10.47$  & $9.90$  & $10.25$  & $9.30$  \\
 & 3: UStyle (w/o $\mathcal{L}_{fft}$ )     & $10.83$  & $9.51$  & $10.67$  & $9.98$  & $11.01$  & $10.60$  & $10.12$  & $9.21$  \\
 & 4: UStyle (w/o $\mathcal{L}_{percept}$) & $10.23$  & $9.30$  & $9.81$  & $9.25$  & $10.76$  & $10.23$  & $9.98$  & $9.09$  \\
 & 5: UStyle (w/o $\mathcal{L}_{clip}$) & $10.15$ & $9.43$  & $9.90$  & $9.40$  & $10.37$  & $10.19$ & $9.75$ & $9.01$ \\
 & 6: UStyle (w/o $L3$)   & $10.23$ & $9.77$  & $9.93$  & $9.45$  & $10.38$  & $10.28$  & $9.70$  & $9.15$  \\
 & 7: \textbf{UStyle (Full)}     & $10.08$  & $9.39$  & $9.89$  & $9.35$  & $10.31$  & $9.50$  & $9.70$  & $8.96$  \\
\bottomrule

\end{tabular}
\end{table*}
\begin{figure}[t]
    \centering
    \includegraphics[width=0.75\textwidth]{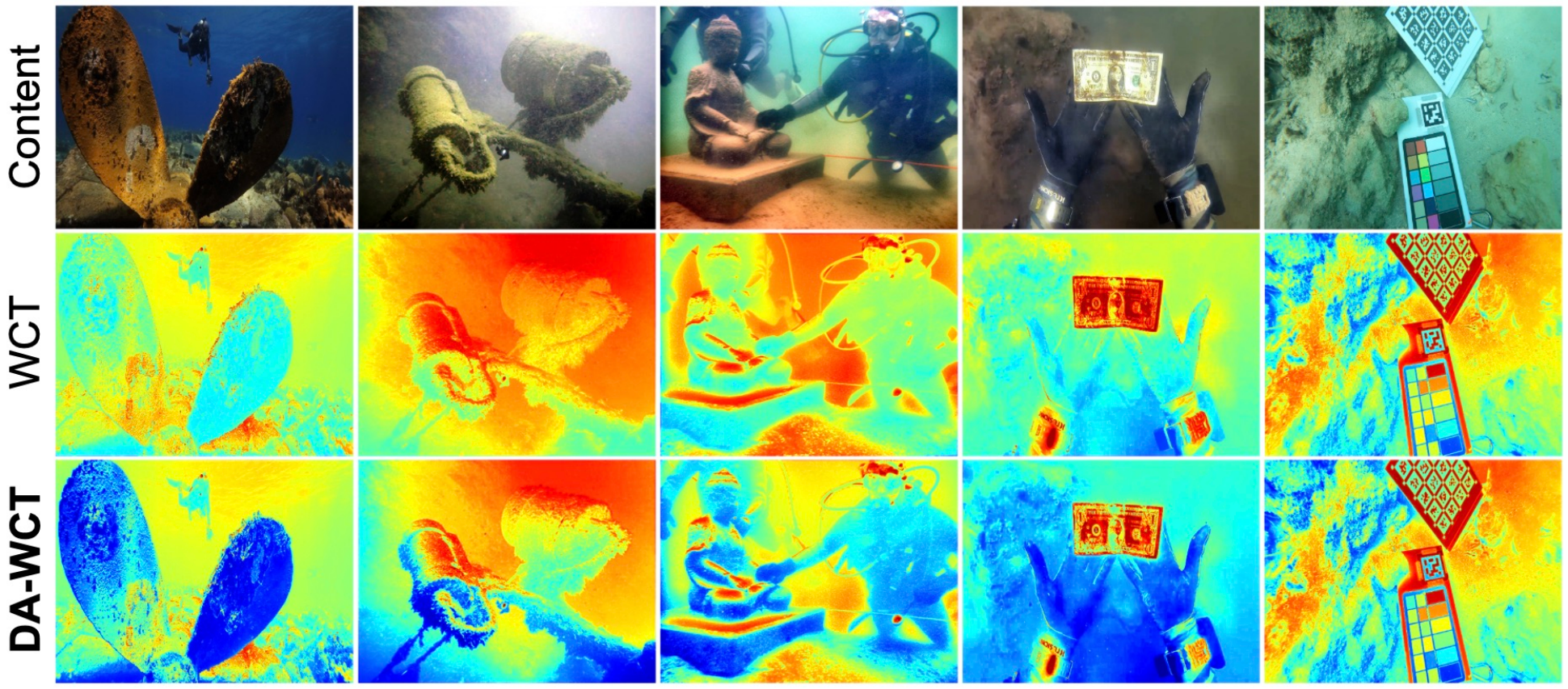}%
    \vspace{-1mm}
    \caption{Grad-CAM heatmap visualizing the effect of depth-aware style fusion in UStyle. Our DA-WCT enhances attention to structural details, ensuring a more spatially coherent style adaptation (third row). In comparison, without depth supervision, attention is unevenly distributed, leading to potential distortions and loss of fine details in complex underwater scenes (second row).}
    \label{fig:gradcam_abl}
\end{figure}

\vspace{-1.0mm}
\subsection{Ablation Studies}\label{ablation}
We conduct ablation experiments to evaluate the key components of UStyle: (\textbf{i}) the impact of depth supervision in the stylization process and (\textbf{ii}) the contribution of various loss components to STx performance. To assess the role of depth supervision, we compare stylized results with and without depth guidance. As shown in Fig.~\ref{fig:gradcam}, DA-WCT improves structural retention and color adaptation in complex underwater scenes. Without depth supervision, stylized images exhibit color distortions and detail loss, whereas DA-WCT preserves content integrity throughout the STx process.

Additionally, we investigate the contribution of different loss components by evaluating seven configurations of our model: (1-5) UStyle without (w/o) each individual loss component:
 $\mathcal{L}_{ssim}$, $ \mathcal{L}_{color}$, $ \mathcal{L}_{fft}$, $ \mathcal{L}_{percept}$ and $\mathcal{L}_{clip}$; (6) UStyle without regularization losses, $L3 = \mathcal{L}_{ssim} + \mathcal{L}_{color} + \mathcal{L}_{fft}$ (w/o $L3$); and (7) the full UStyle model with all loss components. Table~\ref{tab:ablation_table} presents the quantitative comparison across PSNR, SSIM, and RMSE for four underwater style transfer scenarios. The results show that the full UStyle model consistently performs better, demonstrating the effectiveness of depth supervision and loss regularization. From a quantitative perspective, depth supervision leads to a substantial improvement in all three metrics. The inclusion of additional losses further refines the model's ability to balance structural preservation and perceptual realism. The full UStyle model achieves the highest PSNR and SSIM values while maintaining the lowest RMSE across all scenarios.

To further examine the impact of DA-WCT stylization, we employ Grad-CAM~\cite{selvaraju2017grad} visualizations to highlight the regions of interest in our model’s decision-making process. As illustrated in Fig.~\ref{fig:gradcam_abl}, depth supervision enables UStyle to focus more effectively on salient structures, ensuring spatially coherent style adaptation. The middle row displays activation maps generated with depth guidance, where attention is evenly distributed across foreground and background elements, enhancing structure retention. In contrast, the bottom row corresponding to the model without depth supervision reveals inconsistent attention patterns, leading to fine-detail loss and exaggerated distortions. Notably, DA-WCT ensures that complex structures, such as marine organisms and underwater artifacts, are better preserved by aligning style fusion with content depth information. 

These findings confirm that \textbf{depth supervision} is essential for maintaining spatial coherence and content preservation in underwater STx. Furthermore, integrating additional \textbf{regularization losses} significantly enhances perceptual quality, reducing distortions and improving color adaptation.

\begin{table*}[t]
\centering
\caption{Impact of depth noise on UStyle; we compare depth maps from Monodepth2, UDepth, and DepthAnythingV2 on the UF7D test set, reporting average PSNR~($\uparrow$), SSIM~($\uparrow$), and RMSE~($\downarrow$). The results show that more accurate depth maps yield smoother and more consistent stylization.
}
\label{tab:depth_ablation}

\begin{minipage}{0.45\textwidth}
\renewcommand{\arraystretch}{1.2}
  \centering
  \footnotesize
  \begin{tabular}{c|l|c|c}
    \toprule
    \cellcolor{gray!10}{\textbf{(a) Metric}} & \cellcolor{gray!10}{\textbf{Depth Estimator}} & \cellcolor{gray!10}{B $\rightarrow$ G} & \cellcolor{gray!10}{G $\rightarrow$ B} \\
    \hline
    \multirow{3}{*}{{PSNR}} 
    & Monodepth2~\cite{godard2019digging} & 28.30 & 28.44 \\
    & UDepth~\cite{yu2022udepth}     & 28.55 & 28.66 \\
    & DepthAnythingV2~\cite{bhat2023depth} & 28.73 & 28.72 \\
    \hline
    \multirow{3}{*}{{SSIM}}
    & Monodepth2~\cite{godard2019digging} & 0.9411 &  0.8761\\
    & UDepth~\cite{yu2022udepth}     &  0.9503 & 0.8799 \\
    & DepthAnythingV2~\cite{bhat2023depth} & 0.9613 & 0.8818 \\
    \hline
    \multirow{3}{*}{{RMSE}}
    & Monodepth2~\cite{godard2019digging} &  9.80 &  9.65\\
    & UDepth~\cite{yu2022udepth}     &  9.52 &  9.40\\
    & DepthAnythingV2~\cite{bhat2023depth} & 9.39 & 9.35 \\
    \bottomrule
  \end{tabular}
  \vspace{2mm}
  
  \includegraphics[width=0.9\linewidth]{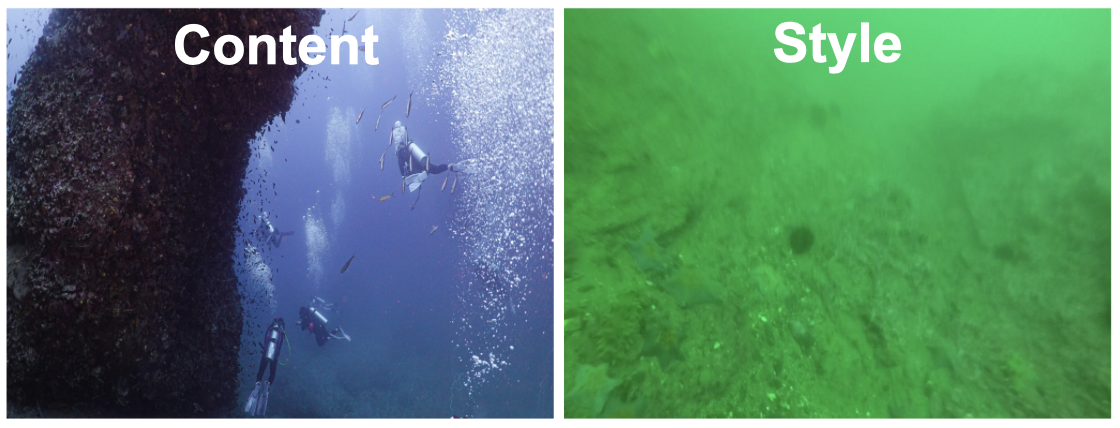}
  \captionof*{figure}{ (b) Content and style images for the comparison shown on the right.}
  \label{Fig:Depth_comparison_style_content}
\end{minipage}
\hfill
\begin{minipage}{0.53\textwidth}
  \centering
  \vspace{-2mm}
  \includegraphics[width=\linewidth]{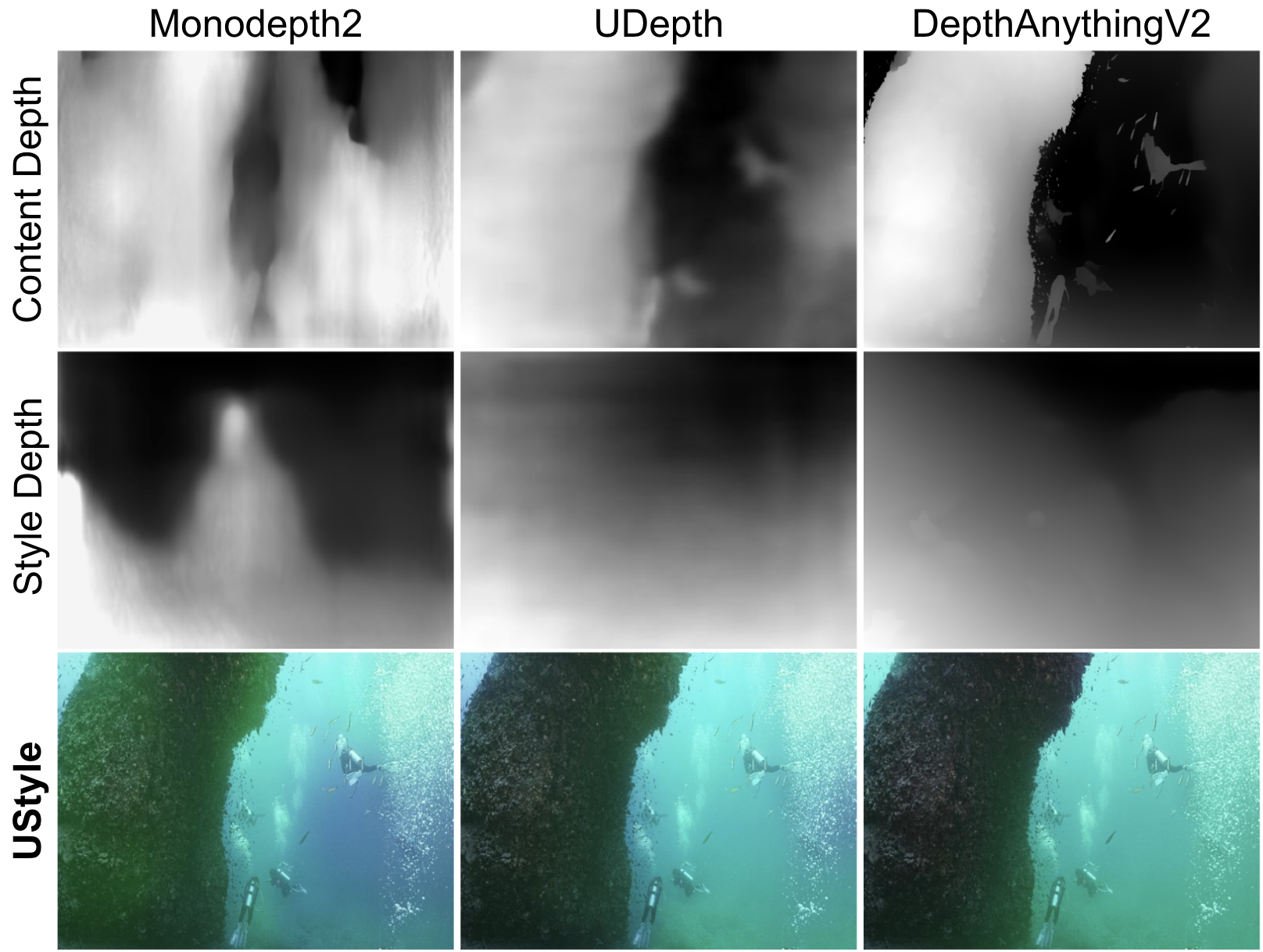}
  \captionof*{figure}{ (c) Stylization with different depth estimators: noisy depth maps from Monodepth2 cause uneven artifacts, UDepth offers moderate improvement, while DepthAnythingV2 generates smoother and coherent stylizations.}
  \label{Fig:Depth_comparison}
\end{minipage}
\end{table*}

\subsection{Analysis of Results} 
\label{subsubsec:analysis_results} 
Our comprehensive evaluation indicates that UStyle offers superior quantitative performance and stable training behavior. We observed that the progressive blockwise training strategy~\cite{wang2018progressive} enabled the network to gradually refine its multi-scale feature representations without significant oscillations in loss values, consistently improving structural fidelity and perceptual quality. As shown in Table~\ref{tab:rmse_psnr_ssim}, UStyle achieves an average RMSE of $8.96$, which is approximately $10$-$16\%$ lower than the values reported by physics-based approaches that range from $10.15$ to $10.45$, respectively, indicating a notable enhancement in reconstruction accuracy. In addition, the perceptual metrics reveal that UStyle attains LPIPS scores as low as $0.1373$ for the DG$\rightarrow$DB transformation and GMSD values around $0.02$, underscoring its effectiveness in preserving edge details and minimizing perceptual distortions. Ablation studies indicate that incorporating depth-aware mechanisms improves SSIM by about $4.5\%$ and reduces RMSE by approximately $10.3\%$. 

Complementary to these numerical results, the Grad-CAM~\cite{selvaraju2017grad} visualizations in Fig.~\ref{fig:gradcam_abl} illustrate that depth guidance enables more uniform and focused attention across critical image regions, ensuring spatial coherence. These findings suggest that the DA-WCT module guides the network to focus on relevant salient structural cues while mitigating artifacts. Moreover, the training process revealed that UStyle consistently converges across diverse underwater conditions without manual parameter tuning, unlike physics-based models, which are sensitive to reference image quality and scene priors. 

During inference using UStyle and the baseline models, we noticed that models without depth guidance, mainly when applied to deep-water style transitions, frequently produced over-saturated outputs with a global averaging effect~\cite{reinhard2001color}; this resulted in increased RMSE values by approximately $10\%$ to $12\%$ and LPIPS scores that were consistently higher. In contrast, UStyle maintained stable convergence and demonstrated a more consistent loss decay pattern with a reduction in loss variance of about $15\%$. Furthermore, the inclusion of the CLIP loss~\cite{radford2021learning} in UStyle contributed to a smoother convergence trajectory by aligning semantic features; removal of this component led to slight degradations in SSIM (a reduction of roughly $0.002$ on average) and a marginal increase in RMSE, highlighting the importance of semantic consistency in waterbody STx learning.

Additional observations from our experiments reveal that UStyle's adaptive multi-stage training strategy~\cite{wang2018progressive} resulted in convergence speeds approximately $20\%$ faster than several baseline models. The fine-tuning process on underwater-specific data proved crucial for the network to adapt to domain-specific distortions, yielding significant improvements in performance metrics across diverse underwater scenarios. This indicates that carefully designed loss functions, particularly those focusing on structural similarity~\cite{hore2010image}, color consistency~\cite{dong2022underwater}, and frequency-domain fidelity~\cite{brigham1988fast}, are essential for achieving high-quality results in complex underwater environments.

Furthermore, as shown in Table~\ref{tab:depth_ablation}, noise in the estimated depth maps and the precision of depth thresholding significantly influence the STx performance in UStyle. Noisy depth maps lead to spatial imbalance in the style transfer, causing uneven blending and reduced retention of fine texture details. In particular, regions with subtle depth gradients are highly sensitive to noise, where Monodepth2 introduces artifacts and inconsistent stylization, while more accurate estimators such as DepthAnythingV2 preserve smooth transitions and maintain structural fidelity in the stylized outputs. Our future research will emphasize robust depth extraction methods and consider dynamic adjustment of depth-based parameters to enhance performance further. Additionally, the success of UStyle in handling diverse underwater conditions without reliance on manually curated reference images with explicit scene priors suggests that its framework may be extended to other domains where scene priors are challenging to obtain.



\vspace{-1.0mm}
\section{Generalization Performance}\label{subsec:discussion}

\vspace{-1.0mm}
\subsection{Unseen Style Transfer}
\begin{figure*}[t]
    \centering
    \includegraphics[width=\textwidth]{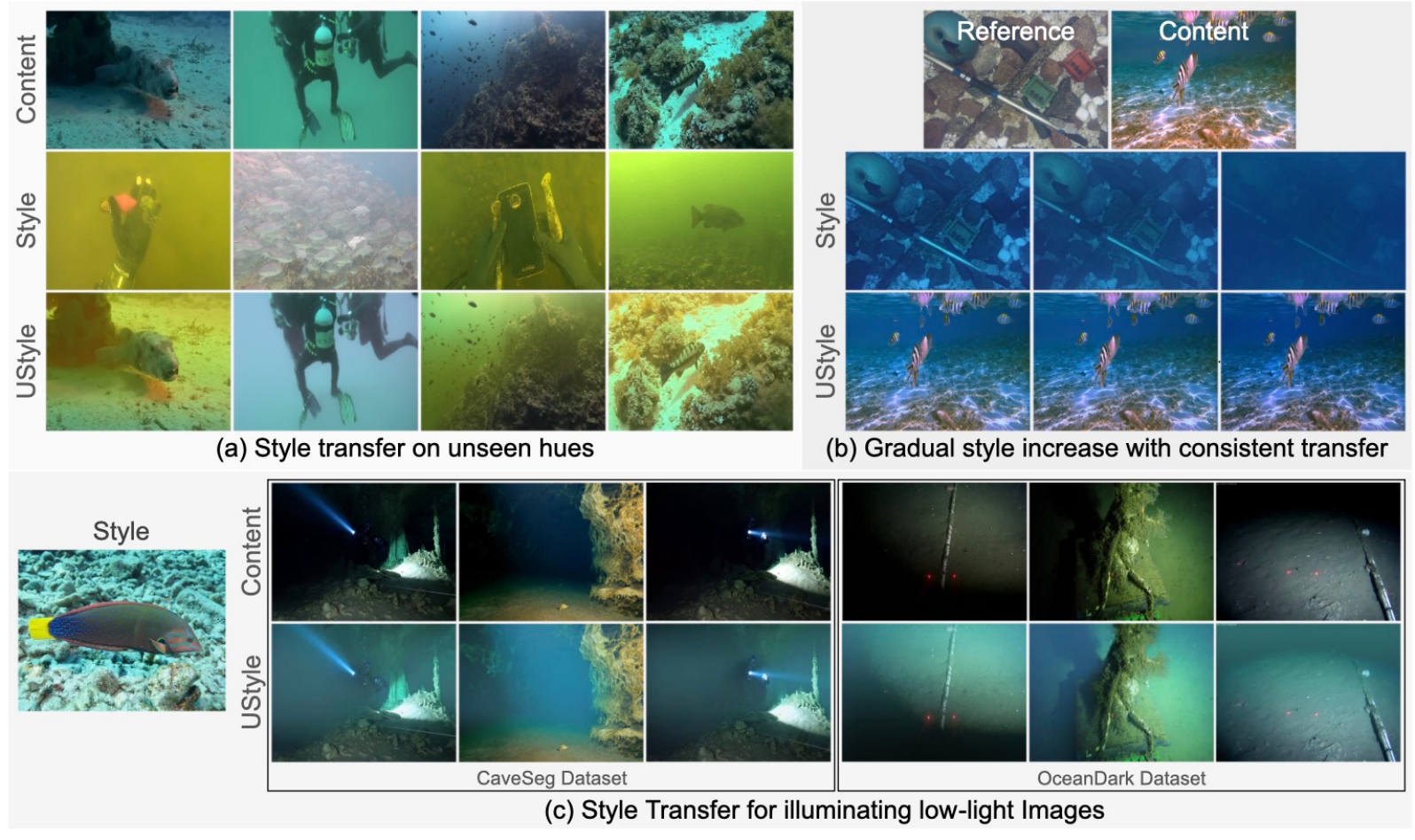}%
    \vspace{-1mm}
    \caption{Evaluation of UStyle's generalization capability under three unseen scenarios: (a) Style transfer on previously unseen color hues (yellow and grey) from UIEB~\cite{UIEB_li}; (b) Consistent interpolation across progressively intensified waterbody styles from TURBID~\cite{duarte2016dataset} dataset; and (c) Illumination of low-light underwater scenes from CaveSeg~\cite{abdullah2023caveseg} and OceanDark~\cite{Oceandark_Marques_2020_CVPR_Workshops} datasets using a well-lit image as style.}
    \label{fig:generalization}
\end{figure*}

{ The generalization capability of UStyle is evaluated under three distinct scenarios as illustrated in Fig.~\ref{fig:generalization}, using the model trained on the UF7D dataset without any additional fine-tuning. In the first scenario, UIEB dataset~\cite{UIEB_li} is used to assess style transfer to unseen waterbody hues, such as grey and yellow, which differ from the predominantly blue and green tones present in the training data. As Fig.~\ref{fig:generalization}(a) shows, UStyle successfully transfers these previously unseen hues onto the content images, demonstrating its strong generalization and robust style representation learned from the UF7D. }
 
{In the second scenario, TURBID dataset~\cite{duarte2016dataset}  is used to examine gradual style intensification. This dataset provides clear underwater images alongside corresponding versions exhibiting progressively increasing turbidity, ranging from light to dark blue tones. When these were used as style references, UStyle achieves smooth and consistent STx across all levels of turbidity, indicating its ability to interpolate within a continuous style manifold. Finally, Fig.~\ref{fig:generalization}(c) highlights the practical applicability under low-light underwater conditions, evaluated using CaveSeg~\cite{abdullah2023caveseg} and OceanDark~\cite{Oceandark_Marques_2020_CVPR_Workshops} datasets. For validation, we apply a well-lit, high-quality reference style image to dimly illuminated underwater scenes. As the results show, UStyle effectively transfers the brighter illumination characteristics, enhancing visibility and preserving structural details. These results demonstrate UStyle's potential for practical underwater image enhancement, beyond stylization alone.}

\begin{figure*}[t]
    \centering
    \includegraphics[width=\textwidth]{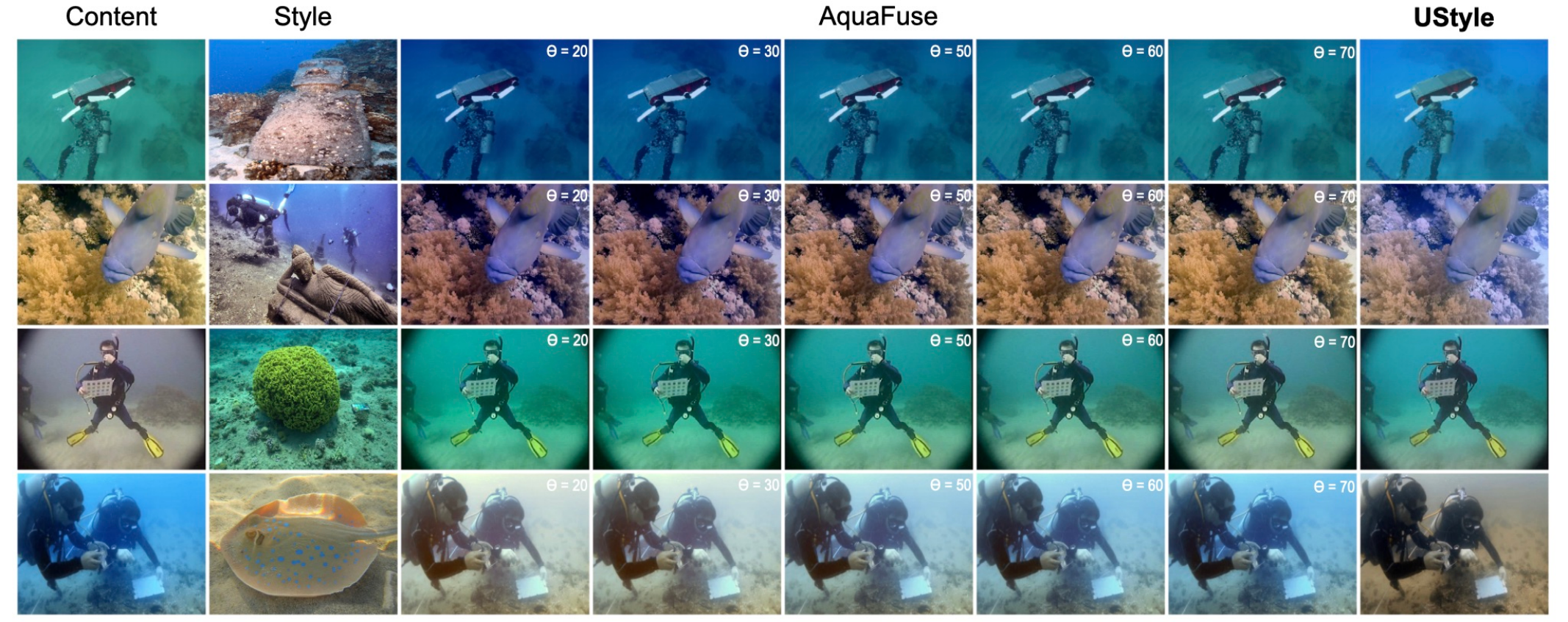}%
    \vspace{-1mm}
    \caption{Comparison between AquaFuse and UStyle across different underwater scenes. The first three rows show that AquaFuse outputs are highly sensitive to the choice of incident angle; the last row further shows that selecting a wrong reference image can significantly impact its waterbody fusion process. UStyle is invariant to these dependencies, offering consistent waterbody STx performance.}
    \label{fig:comparison_aquafuse}
\end{figure*}

\subsection{Benefits Over AquaFuse}
\label{subsubsec:comparison_baselines}
As we discussed earlier in Sec.~\ref{sec:fusion_process}, we use a physics-guided waterbody synthesis module of 
AquaFuse~\cite{siddique2024aquafuse} to extract the waterbody characteristics from the style image. Note that AquaFuse synthesizes the content image by incorporating a reference scene’s waterbody through a physics-guided approximation approach. However, this method relies on a noise-free reference image and prior scene information, including depth, incident angle, and attenuation parameters. As illustrated in Fig.~\ref{fig:comparison_aquafuse}, AquaFuse is highly sensitive to input parameters such as incident angles and tuned attenuation coefficients, limiting its generalizability across diverse underwater conditions. Its performance is strongly influenced by the choice of reference image and variations in incident angles. In contrast, UStyle, a learning-based STx framework, adapts robustly to varying aquatic environments without requiring manually tuned priors. To facilitate this adaptability, large-scale datasets such as USOD10K and UF7D provide a diverse collection of waterbody styles, enabling more generalized STx learning, as demonstrated in our experimental analyses.

\begin{table*}[h]
    \centering
    \caption{Inference speed based on image processing time per image (msec $\downarrow$) across different image resolutions for six data-driven SOTA methods are shown. Evaluations are performed on a single NVidia A100 GPU with $16$\,GB RAM; `$\times$' indicates the out-of-memory (OOM) cases.}
    \label{table_fps_memory}
    \vspace{-1mm}
    \renewcommand{\arraystretch}{1.2}
    \footnotesize
    \begin{tabular}{l r r r r r r r}
    \toprule
       \cellcolor{gray!10}{\textbf{Resolution}} & \cellcolor{gray!10}{$256\times256$} & \cellcolor{gray!10}{$512\times512$} &
       \cellcolor{gray!10}{$640\times480$} &
       \cellcolor{gray!10}{$1024\times1024$} &
       \cellcolor{gray!10}{$1280\times720$} &
       \cellcolor{gray!10}{$1920\times1080$} &
       \cellcolor{gray!10}{$3840\times2160$}\\ 
       \midrule
      ASTMAN~\cite{deng2020arbitrary}        & $180$    & $190.8$    & $222.7$    & $250.6$  & $227.8$ & $274.0$ & $\times$\\
      IEContraAST~\cite{chen2021artistic}    & $1219.5$    & $2127.7$    & $2272.7$    & $2702.7$  & $3125.0$ & $4761.9$ & $\times$\\
      MCCNet~\cite{deng2021arbitrary}        & $170.9$    & $209.2$    & $229.4$    & $354.6$  & $371.7$ & $549.5$ & $1886.8$\\
      StyTr2~\cite{deng2022stytr2}       & $136.2$    & $274.7$    & $285.7$   & $1639.3$  & $2040.8$ & $\times$ & $\times$\\
      PhotoWCT2~\cite{chiu2022photowct2}     & $323.6$    & $340.1$    & $344.8$    & $390.6$  & $374.5$ & $467.3$ & $885.0$\\
      \textbf{UStyle} (ours) & $518.1$  & $621.1$   & $641.0$   & $645.2$  & $645.2$ & $740.7$ & $1087.0$\\
      \bottomrule
    \end{tabular}
    \vspace{-2mm}
\end{table*}

\subsection{Computational Analyses}
As shown in Table~\ref{table_fps_memory}, UStyle maintains a stable inference rate across multiple resolutions. Evaluated on a single NVIDIA A100 GPU, UStyle processes a $256\times256$ image in approximately $518$\,ms and a full HD $1920\times1080$ image in about $741$\,ms. While transformer-based methods such as StyTr2~\cite{deng2022stytr2} and IEContraAST~\cite{chen2021artistic} struggle with high-resolution images due to high memory consumption, UStyle consistently operates across all resolutions, demonstrating scalability. Although MCCNet~\cite{deng2021stytr2} and ASTMAN~\cite{deng2020arbitrary} achieve faster processing times at low resolutions, their performance is not comparable to UStyle. Moreover, due to memory constraints, StyTr2~\cite{deng2022stytr2}, ASTMAN~\cite{deng2020arbitrary}, and IEContraAST~\cite{chen2021artistic} fail to process images at resolutions beyond $1280\times720$ as well. While UStyle is not intended for real-time applications, rather a high-performance data augmentation and stylazation tool, it still provides competitive processing speeds, especially at high input resolutions.

\begin{figure*}[t]
    \centering
    \includegraphics[width=0.95\textwidth]{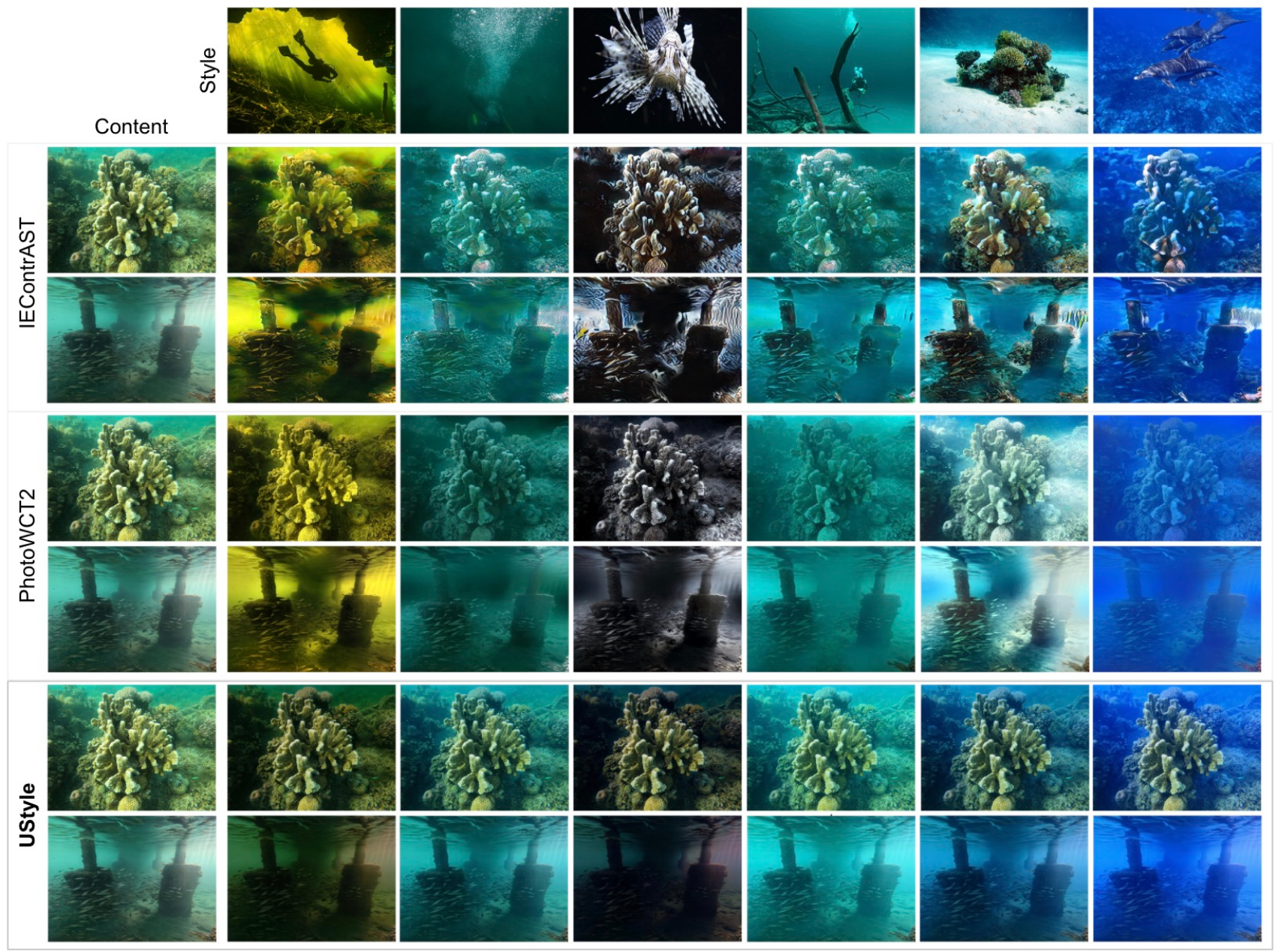}%
    \vspace{-1mm}
    \caption{{Qualitative comparison of underwater STx methods on the UF7D-Challenge dataset. The top row shows different waterbody styles, and each column applies one style to distinct content images. Subsequent rows present outputs from three top-performing models: IEContrAST, PhotoWCT2, and UStyle. As shown, UStyle outperforms the other models and handles the illumination, color tones, and scene structure to generate more realistic underwater stylization.}}
    \label{fig:comparison_challenge}
\end{figure*}
\begin{table*}[t]
\centering
\footnotesize
\caption{{Quantitative comparison of PSNR ($\uparrow$), SSIM ($\uparrow$) and RMSE ($\downarrow$) for UStyle and two best-performing STx methods: IEContrAST and PhotoWCT2. Averaged scores on $100$ ($10\times10$ combinations) for each case in UF7D-Challenge data are reported; best scores are highlighted in \textbf{bold}. }}
\label{tab:challenge_evaluation}
\footnotesize 
\renewcommand{\arraystretch}{1.1} 
\begin{tabular}{l|c|cccccccc}
\toprule
\cellcolor{gray!10}{\textbf{Metric}} & \cellcolor{gray!10}{\textbf{Model}} & \cellcolor{gray!10}{B $\rightarrow$ G} & \cellcolor{gray!10}{G $\rightarrow$ B} & \cellcolor{gray!10}{DG $\rightarrow$ DB} & \cellcolor{gray!10}{DB $\rightarrow$ DG} & \cellcolor{gray!10}{DB $\rightarrow$ C} & \cellcolor{gray!10}{DG $\rightarrow$ C} & \cellcolor{gray!10}{BG $\rightarrow$ C} & \cellcolor{gray!10}{GB $\rightarrow$ C}\\
\midrule
\multirow{3}{*}{{PSNR}}  & IEContrAST & 15.58 & 14.83 & 17.32  & 16.99 & 12.57 & 14.013 & 14.41 & 14.47 \\
& PhotoWCT2 & 16.39 & 16.37 & 18.05  & 18.80 & 13.69 & 13.59 & 15.69 & 14.69 \\
& \textbf{UStyle} (ours) &\textbf{21.21} & \textbf{22.15} & \textbf{25.55}& \textbf{23.31} & \textbf{20.30}& \textbf{20.99}&\textbf{20.47} & \textbf{20.98} \\
\midrule
\multirow{3}{*}{{PSNR}} & IEContrAST & 0.4522 & 0.4829 & 0.5811 &  0.5510 & 0.4066 & 0.4680 & 0.4522 & 0.4718  \\
 & PhotoWCT2 & 0.7479 & 0.7814 & 0.7090 &  0.7454 & 0.6534 & 0.6418 & 0.7641& 0.6990  \\
 & \textbf{UStyle}  (ours) & \textbf{0.8758} & \textbf{0.9047} & \textbf{0.8448} &  \textbf{0.8544} & \textbf{0.8695} & \textbf{0.8922}& \textbf{0.8733}& \textbf{0.8822}  \\   
\midrule           
\multirow{3}{*}{{RMSE}} & IEContrAST & 43.83& 49.24 & 36.45 & 37.87 & 63.67 & 53.71 & 50.16 &  50.40 \\
 & PhotoWCT2 & 43.46& 42.86 & 35.59 & 32.62 & 60.90 & 60.16 & 45.32 &  52.62 \\
 & \textbf{UStyle}  (ours) & \textbf{25.45}& \textbf{23.13} & \textbf{18.30} & \textbf{21.05} & \textbf{28.54} & \textbf{26.91} & \textbf{27.19} &  \textbf{25.84} \\
\bottomrule
\end{tabular}
\end{table*}

\subsection{{Generalization on Uf7D-Challenge}}
\label{subsubsec:Challenge}
{
To further evaluate the generalization performance of UStyle, we curate the \textbf{UF7D-Challenge} test set, which contains $70$ images with $10$ samples for each of the seven waterbody styles in UF7D. The dataset spans a broad range of environments, including open-ocean scenes, underwater caves, and indoor swimming pools, capturing variations in water clarity, color, and structural complexity. It also encompasses diverse illumination conditions: from low-light environments to brightly lit shallow-water scenes -- reflecting natural temporal and lighting variations. The images originate from both online open-access sources and in-field captures using various underwater cameras. Although the exact sensor specifications are unknown, the dataset naturally captures cross-sensor variability due to differences in optics, color calibration, and acquisition settings. Overall, UF7D-Challenge provides a benchmark for evaluating underwater style transfer under heterogeneous environmental and sensor conditions.
}

{
UStyle is evaluated on the challenge subset across eight cross-domain underwater style-transfer scenarios: B$\rightarrow$G, G$\rightarrow$B, DG$\rightarrow$DB, DB$\rightarrow$DG, DB$\rightarrow$C, DG$\rightarrow$C, BG$\rightarrow$C, and GB$\rightarrow$C, comprising a total of $8\times10\times10=800$ test images. 
As illustrated in Fig.~\ref{fig:comparison_challenge}, UStyle produces realistic and perceptually coherent stylizations that preserve geometry and texture across diverse underwater scenes. In contrast, IEContrAST and PhotoWCT2 frequently generate inconsistent stylizations with noticeable content leakage and distorted local structures, particularly under challenging low-light or turbid conditions. UStyle, however, exhibits stable tone adaptation and natural color blending with minimal content leakage, demonstrating its robustness and adaptability for real-world cross-domain underwater style transfer.
}

{
Mroeover, the quantitative results are summarized in Table~\ref{tab:challenge_evaluation}, which demonstrate that UStyle consistently outperforms IEContrAST and PhotoWCT2 across all scenarios, including transfers between highly distinct domains such as G$\rightarrow$B and DG$\rightarrow$C. These results demonstrate strong domain-invariant stylization behavior, with the model effectively adapting to large variations in hue, illumination, and contrast while preserving structural fidelity.
}

\vspace{-2mm}
\subsection{Limitations and Future Directions}

\label{subsubsec:limitations}
Although UStyle demonstrates robust performance and achieves SOTA results, several aspects of the framework warrant further refinement. In particular, enhancements in depth estimation accuracy, training efficiency, adaptive handling of specific waterbody styles, and performance under extreme turbidity conditions remain as challenging cases; more details are listed below.
\begin{enumerate}[label={$\bullet$},nolistsep,leftmargin=*]
\item \textbf{Dependency on depth estimation:} UStyle's performance, as dependent on the DA-WCT module, depends on the quality of depth maps, which can be unreliable in noisy sensing conditions and severely attenuated scenes. Future work could incorporate self-supervised depth estimation methods to enhance robustness and reduce dependency on precomputed scene depth estimation.
\item \textbf{Training and inference efficiency:} Although the progressive blockwise training strategy enhances convergence stability and multi-scale feature learning, it increases training time compared to single-stage models. Future work could explore the design of lightweight architectures or leverage knowledge distillation techniques~\cite{hinton2015distilling} to improve efficiency without sacrificing the generalization performance.
\item \textbf{Extremely turbid or rare waterbody styles:} The compounded effects of scattering and absorption can significantly impair style transfer quality for severely turbid scenes~\cite{ancuti2012enhancing}. techniques. {Future work will explore integrating polarization-based enhancement~\cite{schechner2006recovery} and multi-spectral image enhancement methods~\cite{aggarwal2020fusion} to improve UStyle's robustness in challenging underwater environments. We will also explore adaptive blending mechanisms~\cite{park2019semantic} to better accommodate rare/unseen waterbody styles.}
\end{enumerate}
Despite these challenges, UStyle significantly advances underwater STx literature for data-driven waterbody fusion. Future research will optimize efficiency and extend it to real-time video applications.

\section{CONCLUSION}
\label{sec:conclusion}
In this work, we presented \textbf{UStyle}, the first data-driven underwater waterbody style transfer framework that fuses waterbody characteristics without curated references or explicit scene priors. Leveraging a ResNet-based encoder-decoder, progressive blockwise training, and a novel depth-aware whitening and coloring transform (DA-WCT), UStyle preserves structural integrity while achieving compelling stylization. Extensive evaluations show that UStyle outperforms SOTA artistic, photorealistic, and physics-based methods -- with up to $16\%$ RMSE improvement, PSNR of $29$\,dB, SSIM of $0.96$, and LPIPS as low as $0.13$. Ablation studies confirm that depth supervision and regularization losses enhance geometric consistency and reduce distortions. Moreover, computational analyses reveal that UStyle maintains stable inference rates across various resolutions, showcasing its scalability for high-resolution underwater imaging applications. Although challenges remain in handling extreme turbidity and further optimizing training efficiency, UStyle shows promise for underwater imaging, data augmentation, and vision tasks. Future work will focus on improved depth estimation, lightweight real-time architectures, and extending the framework to more complex scenarios.

\vspace{-2mm}
\section*{ACKNOWLEDGMENT} 
This work is supported in part by the National Science Foundation (NSF) grant \#$2330416$ and the University of Florida (UF) research grant \#$132763$.

\bibliography{refs,robopi_pubs}
\bibliographystyle{IEEEtran}

\end{document}